\documentclass{ieeeaccess}
\usepackage{cite}
\usepackage{amsmath,amssymb,amsfonts}
\usepackage{graphicx}
\usepackage{textcomp}
\usepackage[caption=false,font=footnotesize]{subfig}
\usepackage{dashrule} 
\usepackage{float}
\usepackage{url}
\usepackage{booktabs}
\usepackage{threeparttable}
\usepackage{pdflscape}

\usepackage{algorithm}
\usepackage{algpseudocode}

\usepackage{bm}
\usepackage{newtxtext,newtxmath}

\def\BibTeX{{\rm B\kern-.05em{\sc i\kern-.025em b}\kern-.08em
    T\kern-.1667em\lower.7ex\hbox{E}\kern-.125emX}}

\begin{document}
\doi{}

\title{The Effectiveness of Style Vectors for Steering Large Language Models: A Human Evaluation}
\author{
\uppercase{Diaoulé Diallo}\authorrefmark{1}, 
\uppercase{Katharina Dworatzyk}\authorrefmark{1}, 
\uppercase{Sophie Jentzsch}\authorrefmark{1}, 
\uppercase{Peer Sch\"utt}\authorrefmark{1}, 
\uppercase{Sabine Theis}\authorrefmark{1}, and 
\uppercase{Tobias Hecking}\authorrefmark{1}
}

\address[1]{German Aerospace Center (DLR), Institute of Software Technology, Germany (e-mail: firstname.lastname@dlr.de)}

\markboth
{Diallo \headeretal: The Effectiveness of Style Vectors for Steering Large Language Models}
{Diallo \headeretal: The Effectiveness of Style Vectors for Steering Large Language Models}

\corresp{Corresponding author: Diaoulé Diallo (e-mail: diaoule.diallo@dlr.de).}

\begin{abstract}
Controlling the behavior of large language models (LLMs) at inference time is essential for aligning outputs with human abilities and safety requirements. \emph{Activation steering} provides a lightweight alternative to prompt engineering and fine-tuning by directly modifying internal activations to guide generation. This research advances the literature in three significant directions. First, while previous work demonstrated the technical feasibility of steering emotional tone using automated classifiers, this paper presents the first human evaluation of activation steering concerning the emotional tone of LLM outputs, collecting over 7,000 crowd-sourced ratings from 190 participants via Prolific ($n=190$). These ratings assess both perceived emotional intensity and overall text quality. Second, we find strong alignment between human and model-based quality ratings (mean $r=0.776$, range $0.157$--$0.985$), indicating automatic scoring can proxy perceived quality. Moderate steering strengths ($\lambda \approx 0.15$) reliably amplify target emotions while preserving comprehensibility, with the strongest effects for disgust ($\eta_p^2 = 0.616$) and fear ($\eta_p^2 = 0.540$), and minimal effects for surprise ($\eta_p^2 = 0.042$). Finally, upgrading from Alpaca to LlaMA-3 yielded more consistent steering with significant effects across emotions and strengths (all $p < 0.001$). Inter-rater reliability was high (ICC $= 0.71$--$0.87$), underscoring the robustness of the findings.
These findings support activation-based control as a scalable method for steering LLM behavior across affective dimensions.




\end{abstract}



\maketitle

\section{Introduction}
\PARstart{A}{dvances} in Generative Artificial Intelligence (AI) and the rapid evolution of Large Language Models (LLMs) have ushered in a new era of conversational systems. Their capacity to engage in near-human conversations is increasingly impressive, and they have the potential to be applied across diverse contexts---ranging from education, customer support, and healthcare to safety-critical domains such as aerospace.


With this growing capability comes a pressing need to understand and control their internal processes to ensure consistent and desirable outputs. While LLMs like LlaMA or ChatGPT have showcased remarkable proficiency in language comprehension and generation, much of this behavior remains a black box, particularly when capturing and modulating implicit cues, such as sentiment and emotion.

It has been argued that consistent language generation and delivery of information alone are insufficient for seamless and effective communication between humans and machines. AI agents need to incorporate human factors into language generation and be adaptive concerning the human counterpart to meet the communication goals~\cite{hovy2021social}.

Today, there is a growing body of research exploring methodologies for nuanced steering of style and tone of LLM outputs by manipulating their internal activations during a forward pass~\cite{konen2024style, turner2023activation, rimsky2024steering, vanderweij2024extending, wang2024sadi, soo2025fgaa, silva2025steering}. These approaches are based on the assumption that concepts such as sentiments have a representation in the activation vector space of LLMs, and thus, the expression of these concepts can be influenced by vector operations in this space. Therefore, they are sometimes subsumed as activation steering. It could be shown that activation steering is a resource-efficient technique that can influence the tone of an LLM in a very nuanced and controllable fashion \cite{konen2024style}, which is difficult to achieve with prompt engineering. 

Following the stylized response generation literature \cite{sun2022expanding,yang2021fudge,gao2019structuring,jin2020hooks} and our prior work \cite{konen2024style}, we use \emph{style} as an abstract, content-agnostic way of expression—how something is said rather than what is said.
Typical style attributes include emotion (e.g., joy, anger), sentiment (positive/negative), politeness, formality, or persona. In this work, we instantiate style as affective tone.

For intuition, think of style as a lightweight “filter” laid over the same content: the facts stay the same, but the phrasing shifts toward a chosen manner of expression (e.g., more joyful or more formal).


This paper advances this research in three key aspects: So far, all works on activation steering lack a human evaluation. The possibility of steering the tone of LLMs has been demonstrated only through automatic classification of outputs or illustrative examples. Whether humans can perceive nuanced steering in the style of LLMs is an open question, which is crucial for moving activation steering from experimentation to real human-machine interaction applications. This evaluation is critical as research has shown that human perception of AI systems—especially emotional dimensions—significantly impacts trust, acceptance, and adoption intention, explaining up to 78.5\% of variance in users' willingness to use AI in safety-critical environments \cite{schon2025cleared}. To this end, we present the first human evaluation of activation steering, involving over 7,000 crowd-sourced judgments on perception of emotional tone and text quality. 

Second, we analyze the relationship between steering strength and output quality, showing that moderate interventions preserve fluency while effectively shifting emotional expression. We also find that human and automatic quality ratings are closely aligned, indicating the potential of model-based scoring as a proxy in future evaluations.

Furthermore, we adapt the method of \cite{konen2024style} to the more capable LlaMA‑3 architecture, which exhibits more consistent and pronounced steering effects. 

The remainder of this paper is structured as follows: Section~\ref{sec:related_work} reviews related work on model steering and latent intervention methods. Section~\ref{sec:methodology} outlines our activation steering approach. Section~\ref{sec:experiments} describes our experimental setup, including model configurations, datasets, and automatic evaluation metrics. Section~\ref{sec:human_evaluation} details the design and execution of our human evaluation. Section~\ref{sec:results_steering} presents the results, including analyses of steering effectiveness, alignment between human and model evaluations, and the impact on text quality. Finally, we conclude with a discussion of limitations and directions for future work in Section~\ref{sec:conclusion}.

\section{Related Work}
\label{sec:related_work}

The introduction of transformer architectures~\cite{vaswani2017attention} significantly advanced the capabilities of LLMs, enabling models such as GPT-3~\cite{brown2020language} and GPT-4~\cite{openai2023gpt4} to encode abstract concepts implicitly, including stylistic features and emotions~\cite{schramowski2022large, wagner2022does}. Various methods have been proposed to effectively control or steer these outputs.

Prompt engineering directly influences model outputs by crafting specific textual prompts~\cite{keskar2019ctrl, shin2020autoprompt, lester2021power, li2021prefix, wei2022chain}. Although successful in task-specific scenarios, it often requires extensive manual optimization and lacks precise stylistic control. Traditional fine-tuning methods in stylized response generation~\cite{gao2019structuring, jin2020hooks, yang2021fudge, sun2022expanding} offer deeper stylistic adjustment but involve significant computational resources, making them impractical for large-scale models. 

Activation engineering modifies internal activations of LLMs during inference without retraining. Subramani et al.~\cite{subramani2022acl} initially introduced steering vectors optimized for generating predefined sentences from empty inputs, which required costly per-sentence optimization.
Turner et al.~\cite{turner2023activation} advanced this by computing steering vectors from differences between semantically opposite prompt pairs (e.g., "love" versus "hate"). These vectors directly altered model activations to effectively steer outputs in a targeted sentiment direction.
Konen et al.~\cite{konen2024style} introduced activation-based style vectors, aggregating activations from labeled datasets representing various style categories. By contrasting the aggregated activations of a target style against the mean activations from contrasting styles, our approach produced generalized, computationally efficient vectors without requiring paired examples.
Subsequently, Rimsky et al.~\cite{rimsky2024steering} independently proposed Contrastive Activation Addition (CAA). Their method computed steering vectors by averaging activation differences across large sets of positive and negative examples, injecting these vectors at every token position for the control of behaviors such as factuality and sycophancy.
Other studies have expanded the scope of steering multiple simultaneous behaviors. For example, van der Weij et al.~\cite{vanderweij2024extending} combined multiple steering vectors, observing significant interference effects and sensitivity to layer selection. 

Beyond steering, recent work probes human--AI emotional alignment. Systematic reviews report that LLM empathy is often appropriate yet imperfectly aligned with human references~\cite{valluripalli2025}, and human studies find model outputs can be perceived as empathic depending on presentation and attribution~\cite{kraus2025,lee2025}. Our human evaluation complements these findings by isolating graded, activation-space control and measuring perceived intensity/comprehensibility.

As outlined above, prompt-based methods are fragile and lack a scalar control of intensity, while fine-tuning variants improve control at the cost of training and deployment overhead. Activation-based style vectors complement this space by providing: (i) graded, high-precision control via a single parameter~$\lambda$; (ii) no re-training, since vectors are computed once from labeled corpora; and (iii) the ability to capture and reproduce stylistic differences present in a corpus—even when these differences are difficult to verbalize explicitly—by operating directly in activation space rather than through hand-crafted textual descriptions.



Evaluation methods for activation steering predominantly rely on automated metrics, such as classifier-based sentiment analysis or GPT-based proxies~\cite{turner2023activation, rimsky2024steering, konen2024style}. The broader literature lacks rigorous, large-scale human-based assessments concerning nuanced emotional or stylistic steering. Several studies note that steering strength can negatively affect output quality, especially in reduced fluency or increased incoherence at higher steering intensities~\cite{silva2025steering, rimsky2024steering, dathathri2019plug, yang2021fudge}. The present study addresses this research gap by combining human-centered evaluations with automated classifier metrics to assess the perceptual validity and interpretability of steering effects. Additionally, it investigates the ecological validity and mental model alignment of affective steering techniques in large language model outputs.

\section{Methodology}\label{sec:methodology}
In this section, we will first briefly describe the activation steering pipeline based on~\cite{konen2024style}. Subsequently, we comprehensively elaborate improvements, as a special focus of this work lies on the updated experimental pipeline.

\begin{figure}
    \centering
    \includegraphics[width=\linewidth]{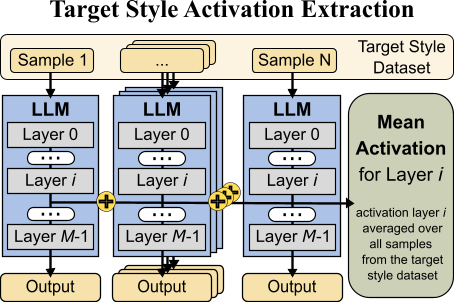}
    \caption{Illustration of target style activation extraction: N samples from a target style are fed into the LLM, and the mean activation at each layer $i$ is computed to represent the characteristic activation pattern of that style.}
    \label{fig:getting_av}
\end{figure}

\begin{figure}
  \centering
  \includegraphics[width=\linewidth]{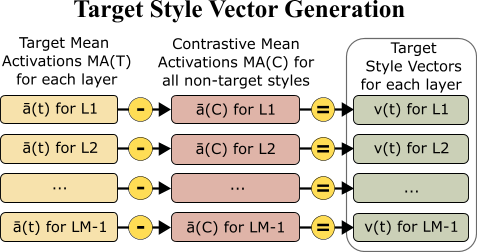}
  \caption{%
    Computation of style vectors: for each layer~$i$, the style vector
    $\mathbf v^{(i)}(t)$ is obtained by subtracting the contrastive mean
    activation
    $\overline{\mathbf a}^{(i)}(\mathcal C)
      = \frac{1}{s}\sum_{j=1}^s
        \overline{\mathbf a}^{(i)}(c_j)$
    from the target mean activation
    $\overline{\mathbf a}^{(i)}(t)$.%
  }
  \label{fig:av_to_sv}
\end{figure}

We aim to modify an LLM's responses to any input prompt $x$ during an ongoing conversation by steering them toward a specific target style $t$. To this end, a set of layer-wise style vectors for a predefined target style $t$ is constructed based on a labeled dataset.

We define the set of style categories as $\mathcal{Y} = \{\,t,\,c_1,\dots,c_s\}$, where $t$ denotes the \emph{target} style and $c_1, \dots, c_s$ are $s$ contrastive (non-target) styles. For each label $y \in \mathcal{Y}$, we assume a set of representative samples $d^{(n)}_y$ from a labeled corpus. Each sample is passed through the LLM in a forward pass to collect hidden activations from each layer $L_i$ (see Fig.~\ref{fig:getting_av}).

The per-layer activations are then averaged over all samples of a specific style label. We denote the mean activation vector at layer $L_i$ for the target style $t$ as $\bar{\mathbf a}^{(i)}(t)$ and for contrast style $c_j$ as $\bar{\mathbf a}^{(i)}(c_j)$. 



To capture what distinguishes the target style from the others, we compute the style vector $\mathbf{v}^{(i)}(t)$ for layer $L_i$ as the difference between the target and contrastive mean activations:

\begin{equation}
\mathbf{v}^{(i)}(t)
= \bar{\mathbf{a}}^{(i)}(t)
  \;-\;
  \frac{1}{s}
  \sum_{j=1}^{s}
    \bar{\mathbf{a}}^{(i)}(c_j)
\label{eq:style_vector_inline}
\end{equation}

This vector encodes a direction in activation space that characterizes the unique traits of the target style $t$ at each layer (see Fig.~\ref{fig:av_to_sv}).

\begin{figure}
    \centering
    \includegraphics[width=\linewidth]{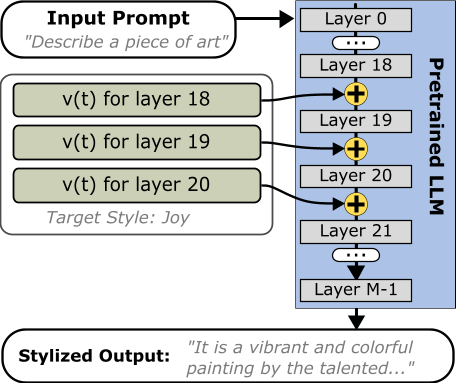}
    \caption{Process of steering model activations during inference. In this illustration, only three layers—18, 19, and 20—are modified using activation-based style vectors for the target style joy; in our full experiments, we inject the vectors into all model layers.}
    \label{fig:steering_process}
\end{figure}

At inference time, the steering is performed by modifying the original activations $\mathbf{a}^{(i)}(x)$ at each layer $L_i$ using the corresponding style vector (see Fig.~\ref{fig:steering_process}). The updated activations become:

\begin{equation}
\mathbf{\hat{a}}^{(i)}(x) = \mathbf{a}^{(i)}(x) + \lambda \cdot \mathbf{v}^{(i)}(t)
\label{eq:inference}
\end{equation}

Here, $\lambda$ is a scaling parameter that controls the intensity of the steering effect and is shared across layers. Operationally, $\lambda$ behaves like a single fader for the target emotion—pushing it higher intensifies the tone, pulling it lower restores a neutral baseline.
This method enables nuanced and controllable stylistic modulation during generation, offering an alternative to prompt-based conditioning.
Algorithm~\ref{alg:style_steering} summarizes vector construction (Eq.\ref{eq:style_vector_inline}) and inference-time integration (Eq.\ref{eq:inference}) as pseudocode.

\begin{algorithm}[t]
\caption{Activation-based Style Steering (construction and inference)}
\label{alg:style_steering}
\begin{algorithmic}[1]
\Require Labeled corpus $\mathcal{D}$; style labels $\mathcal{Y}=\{t,c_1,\dots,c_s\}$ with $\mathcal C=\{c_1,\dots,c_s\}$; LLM layers $L_0,\dots,L_{M-1}$; steering strength $\lambda$
\Ensure Per-layer style vectors $\{\mathbf v^{(i)}(t)\}_{i=0}^{M-1}$ and steered continuation of $x$

\Procedure{BuildStyleVectors}{$\mathcal{D}, \mathcal{Y}$}
  \For{$i \gets 0$ \textbf{to} $M-1$}
     \State $\overline{\mathbf a}^{(i)}(t) \gets \text{mean activation at } L_i \text{ over samples of } t$
     \State $\overline{\mathbf a}^{(i)}(\mathcal C) \gets \frac{1}{s}\sum_{j=1}^s \overline{\mathbf a}^{(i)}(c_j)$
     \State $\mathbf v^{(i)}(t) \gets \overline{\mathbf a}^{(i)}(t) - \overline{\mathbf a}^{(i)}(\mathcal C)$ \Comment{Eq.~\ref{eq:style_vector_inline}}
  \EndFor
  \State \Return $\{\mathbf v^{(i)}(t)\}_{i=0}^{M-1}$
\EndProcedure

\Procedure{SteerAtInference}{$x$, $\{\mathbf v^{(i)}(t)\}$, $\lambda$}
  \For{token step $u=1,\dots,U$}
     \For{$i \gets 0$ \textbf{to} $M-1$} \Comment{steer all layers; see Fig.~\ref{fig:steering_process}}
        \State $\hat{\mathbf a}^{(i)}(x) \gets \mathbf a^{(i)}(x) + \lambda \cdot \mathbf v^{(i)}(t)$ \Comment{Eq.~\ref{eq:inference}}
     \EndFor
     \State Generate next token using steered activations across all layers
  \EndFor
\EndProcedure
\end{algorithmic}
\end{algorithm}

In our study, style concepts refer to emotion categories such as \textit{joy}, \textit{anger}, or \textit{disgust}, but this approach generalizes to any style distinguishable via a representative corpus.

\section{Experiments}
\label{sec:experiments}
This section provides a detailed overview of the experimental setup used to evaluate the effectiveness of activation-based emotion steering in LLMs. Our overall approach involves three main components: (1) constructing style vectors in the activation space of the LLM corresponding to different target emotions, (2) prompting the model with input texts and systematically applying these style vectors at varying intensities, and (3) evaluating the resulting outputs through both automatic classification models and human assessment.

Through this pipeline, we investigate whether steering the internal activations of a language model leads to perceivable and interpretable changes in the emotional tone of its outputs. We first describe the selected LLM and its configuration, followed by the dataset used to derive emotion-specific style vectors. We then present our automatic evaluation setup and, finally, the design and implementation of a human evaluation to assess perceived emotional intensity and text quality.

\subsection{Models and Parameters}
We initially experimented with several variants of instruction-tuned language models and selected Llama-3-8B-Lexi-Uncensored~\cite{llama3lexi} due to its superior performance in preliminary style vector evaluations. Compared to the standard LlaMA-3 model, this variant demonstrated better control capabilities in latent activation space, especially in tasks involving emotional and stylistic conditioning.

To extract hidden activations from all model layers during forward passes, we processed textual samples from our dataset with a maximum token length of 300. This constraint was necessary to avoid GPU memory overflows during batch processing. If input prompts exceeded this token threshold, we iteratively truncated the input text from the end until it satisfied the limit. Although this allowed stable processing, it occasionally resulted in incomplete or semantically truncated sentences.

\subsection{Dataset}
GoEmotions~\cite{demszky2020goemotions} is a multi-class style dataset consisting of $58k$ manually curated user comments from the internet platform Reddit. Comments are labeled with 27 emotional categories. From this, we extract a subset of $53,994$ unique samples that can be unambiguously mapped to one of the six basic emotion categories defined by Ekman~\cite{ekman1971}: \emph{sadness}, \emph{joy}, \emph{fear}, \emph{anger}, \emph{surprise}, and \emph{disgust}. The mapping from the original 27 emotion labels to the six Ekman categories was performed using the official mapping provided in the GoEmotions repository~\cite{goemotions_mapping}. The distribution of unique samples per Ekman emotion label is shown in Table~\ref{tab:emotion-distribution}.

\begin{table}[!ht]
\caption{\textbf{Sample distribution per emotion label after Ekman mapping}.}
\label{tab:emotion-distribution}
\setlength{\tabcolsep}{8pt}
\centering
\begin{tabular}{l r}
\hline
\textbf{Emotion} & \textbf{Sample Count} \\
\hline
Joy        & 19,440 \\
Neutral    & 17,716 \\
Surprise   & 5,839 \\
Anger      & 5,682 \\
Sadness    & 3,622 \\
Disgust    &   881 \\
Fear       &   814 \\
\hline
\end{tabular}
\end{table}

Although the \emph{neutral} category is not part of Ekman's six basic emotions, it was included in the construction of style vectors to capture general-purpose directions in the model's latent space. 

\subsection{Automatic Evaluation}
To complement the human ratings, we automatically evaluated using two model-based scorers: one for emotion intensity and another for text comprehensibility. For emotion classification, we employed a pre-trained transformer model fine-tuned on English emotion detection, specifically DistilRoBERTa ~\cite{hartmann2022emotionenglish}. For this model an overall classification accuracy of 66\% on a balanced six-emotion evaluation set (chance level $\approx 14\%$) was reported. The model outputs a probability distribution across emotion categories, which we use to track how strongly the target emotion is expressed in the generated text. 

We applied the same LLM used for text generation in a scoring mode to evaluate comprehensibility. The model was prompted to rate each sentence on a scale from 1 (highly comprehensible) to 10 (incomprehensible), based on predefined instructions emphasizing clarity and logical coherence.

\section{Human Evaluation}
\label{sec:human_evaluation}
\subsection{Background}
\label{subsec:theoretical_background}
Paul Ekman's model~\cite{ekman1971, ekman1992} encompasses the six basic emotions of \textit{anger}, \textit{disgust}, \textit{fear}, \textit{joy}, \textit{sadness}, and \textit{surprise} and justifies its application by recognizing such emotions as universal and cross-cultural. 
His studies showed that these emotions are displayed and identified similarly around the world, indicating their potential suitability for applications such as emotion control — a possibility further explored in the present study.
Another motivation for adopting this emotion model in the present study was that most available training datasets are based on the Ekman model. The distinction between the corresponding emotion categories supports our assumption that people with an unrestricted ability to perceive emotions can recognize the differences in intensity within these categories. A study on cognitive appraisal methods showed humans can identify emotion in texts with the help of contextual cues even if these are implicit (e.g., meeting a snake is associated with fear)~\cite{hofmann2020appraisal}. However, Ekman's primary emotions are derived from studies of facial expressions. There is currently little knowledge about the human perception of emotions in text. Initial approaches focus primarily on developing annotation schemes and their validation~\cite{troiano2023emotions}.  

\subsection{Sample}
\label{subsec:sample}
In total, 227 participants (50\% male and 50\% female) were recruited via Prolific~\cite{prolific2024}. To account for sex-specific differences in emotion interpretation, assignment to the different prompt conditions was balanced for participants’ sex. Each of the 19 prompt conditions was completed by 5 female participants and 5 male participants. As the study involved extensive reading and understanding of English texts, participants were selected if their first language was English and they had no literacy difficulties. Responses from participants were excluded if they did not pass the prescreening (see Section ~\ref{subsubsec:emotion_detection_ability}), indicating insufficient emotion detection abilities, or if a comparatively short completion time indicated low-effort responses. Exclusion of 37 responses yielded a final sample of 190. Of the final sample, 50\% reported their gender as woman, 49.5\% as man, and one person as non-binary. The average age was 38.42 (\textit{SD} = 13.57) with a range of 18 to 78 years. All participants reported English as their first language and no literacy difficulties; all but one rated their reading fluency as \textit{completely fluent} (88.4\%) or \textit{very fluent} (11.1\%). The majority was currently residing in the United Kingdom (52\%), followed by Canada (15\%), South Africa (13\%), United States (12\%), Ireland, New Zealand, Australia, Spain, and Korea (less than 5\%). A total of 62\% identified with Western or European culture, 28\% with African culture, 4\% with Asian culture and the remaining 6\% with another culture (e.g. Middle Eastern or Indigenous). Most of the participants were highly educated (Bachelor’s degree: 46.9\%, Master’s degree: 20.0\%, Doctoral degree: 3.2\%), full-time or part-time employed (89\%), and comfortable in interpreting emotions in English text (\textit{extremely comfortable}: 62\%, \textit{very comfortable}: 35\%, \textit{moderately comfortable}: 3\%).
An a priori power analysis (F-tests, repeated measures ANOVA) using G*Power determined a sample size of \textit{n} = 190 to be adequate to detect a small effect with 95\% power at a significance level of \(\alpha\) = .05.

\subsection{Study Design}
\label{subsec:design}
To investigate how humans perceive the emotion steering strength of an LLM's text output when steered to six basic emotion types, a 6 × 8 within-subjects design was employed. Therein, we considered the effect of the independent variables \textit{target emotion} (\textit{anger}, \textit{disgust}, \textit{fear}, \textit{joy}, \textit{sadness} and \textit{surprise}) and \textit{steering strength (\(\lambda\))} (0.00, 0.05, 0.1, 0.15, 0.2, 0.25, 0.3, 0.35) on the dependent variables \textit{perceived emotional intensity} and \textit{text comprehensibility}.  
As there is currently little empirical evidence regarding the perceptibility of emotional control intensity in texts, we have decided to initially look at very small-step adjustments of the \(\lambda\) value in the present experiment.
Furthermore, high \(\lambda\) values affect the generated texts' coherence, readability, and naturalness. Preliminary experiments revealed that steering strengths beyond $\lambda=0.35$ frequently led to heavily degraded semantic coherence and overall output quality, and were excluded from the final evaluation.

\subsection{Materials and Measures}
\label{subsec:measures}

\subsubsection{Stimuli}
\label{subsubsec:stimuli}
Texts were generated by the LLM using predefined prompts for each target emotion and \(\lambda\) value (for examples see Appendix~\ref{app:example}). The generated texts were identical across all target emotions for \(\lambda\) = 0.00.

\subsubsection{Emotion Detection Ability}
\label{subsubsec:emotion_detection_ability}
Prescreening included a self-administered test on emotion detection ability in text, ensuring the inclusion of participants with adequate emotion recognition skills. As part of the test, participants rated six texts from the International Survey on Emotion Antecedents and Reactions (ISEAR) dataset collected by the Swiss Center for Affective Sciences; the dataset comprises 7,666 self-reported emotional experiences from nearly 3,000 individuals across diverse cultural backgrounds \cite{scherer1994evidence}. Each entry in ISEAR includes a brief textual description of an emotional event, categorized into seven primary emotions—\textit{joy}, \textit{fear}, \textit{anger}, \textit{sadness}, \textit{disgust}, \textit{shame}, and \textit{guilt}—based on appraisal theories of emotion. ISEAR has been extensively used in machine learning models for emotion classification, providing a well-validated benchmark for lexical, syntactic, and contextual emotion cues\cite{bostan2018analysis}. For each emotion category in Ekman's model represented in ISEAR, we selected sentence IDs (\textit{joy}: 444, \textit{fear}: 214, \textit{sadness}: 467, \textit{disgust}: 124, \textit{anger}: 42), with a large word count to better match the length of our LLM output texts.
Unlike traditional emotion recognition scales that rely on categorical labels or forced-choice paradigms, the proposed method employs a continuous rating system where participants assess the intensity of six emotions (\textit{joy}, \textit{anger}, \textit{sadness}, \textit{fear}, \textit{disgust}, and \textit{surprise}) within a 0--7 scale for various textual scenarios. Dominant emotion for each text was determined by the highest-rated emotion with a flexible threshold ($\pm0.5$ around the maximum value) to account for natural variability in human perception, using a scale adjustable in 0.1 increments. When multiple emotions were rated highly relevant, reflecting the reality of human affective processing, participants must correctly identify the dominant emotion in at least three out of five passages to pass the screening. This ensures that minor variations or ties do not lead to unnecessary exclusions. Compared to standardized emotion recognition tests, such as Ekman’s Facial Emotion Recognition Task or the Reading the Mind in the Eyes Test, this method provides superior ecological validity by evaluating emotion perception in context rather than relying on static, artificial stimuli. Furthermore, the continuous scale mitigates floor and ceiling effects, offering a granular assessment of emotion perception accuracy. By integrating context-sensitive, flexible, and adaptive scoring criteria, this pre-screening method better reflects real-world emotion recognition, ensuring that subsequent survey responses are derived from individuals with sufficient emotion detection ability.

\subsubsection{Demographics}
\label{subsubsec:demographics}
Participants were asked to indicate their gender, age, country of residence, culture, level of education, field of study or work, first language, level of reading fluency, and emotion interpretation. Additional demographic data, such as sex, literacy difficulties, and employment status, were obtained from their Prolific profile.

\subsubsection{Emotional Intensity}
\label{subsubsec:emotional_intensity}
The emotional intensity participants perceived when reading the generated texts was assessed using six rating scales. Each rating scale represented one of the six basic emotions and ranged between 0 and 7 with 0.1 increments, allowing for a fine-grained assessment. The same response format as in the prescreening was used to be able to accurately represent the complexity of human perception. According to Ekman, such rating scales allowed the participants to indicate the intensity of the basic emotions they associate with the text. 
In formulating the instructions, the emphasis was placed on evaluating the emotion conveyed by the text and not on the emotion elicited in the participant (“Select the intensity of each emotion and the comprehensibility you would assign to the following text.”). A score of 0 indicated that the person could not associate any intensity of emotion with the text. In contrast, a score of 7 showed that the emotion was associated with the text at the highest possible intensity.

\subsubsection{Text Comprehensibility}
\label{subsubsec:text_comprehensibility}
To assess the overall text quality, participants rated the comprehensibility of the output texts ("Select the intensity of each emotion and the comprehensibility you would assign to the following text") on the same 0 to 7 scale as emotional intensity, where 0 indicated low comprehensibility and 7 high comprehensibility.

\subsection{Procedure}
\label{subsec:procedure}
The study was transparently introduced as an investigation of human perception of emotions in written texts. Participants were informed about the scope and the structure of the study, including the prescreening, and the risk of experiencing discomfort when reflecting on emotions or experiences. In addition to providing information about the use of the data, participants were assured that all responses were collected anonymously and encouraged to take breaks during the study.
After providing informed consent via the survey tool LimeSurvey~\cite{limesurvey2024}, participants started the experiment by rating five prescreening texts. Participants who failed to correctly rate at least three prescreening texts were redirected to the Prolific site and did not participate in the main study. Participants who completed the prescreening successfully were asked to answer ten demographic questions. Next, participants were randomly assigned to one of 19 prompts (see Appendix \ref{app:prompts}). Texts were generated for the six basic emotions and eight steering strengths for each prompt, resulting in 42 distinct and 6 identical (\(\lambda\) = 0.00) outputs. Participants rated each text on perceived emotional intensity and text comprehensibility (``Select the intensity of each emotion and the comprehensibility you would assign to the following text.''). The order in which participants rated the texts was randomized with respect to target emotion, \(\lambda\) value, and perceived emotional intensity. 
At the end of the study, participants were redirected to the Prolific site and received a completion code. After manually reviewing the submitted responses, participants who completed the research received compensation of \pounds7.50. Participants who were screened out and did not participate in the central part of the study received compensation of \pounds0.10 per minute. The study duration was estimated at 60 minutes; the actual duration ranged between 14 and 156 minutes, with an average of 65 minutes (\textit{SD} = 29). Throughout the study, participants could contact the experimenters through the Prolific platform. Participants reported two technical issues (incorrect display of the progress bar and failed randomization of one output text), which were subsequently resolved by the experimenters.

\subsection{Statistical Analysis}
\label{statistical_analysis}
Missing data for the target emotion \textit{joy} and \(\lambda\)-value 0.10 and 0.25, which occurred due to a malfunction in the randomization process in nine cases, were replaced with mean values (54 replacements in total, < 1\%). We then conducted a two-way repeated-measures analysis of variance (ANOVA) to test for differences in the perceived emotional intensity as well as text comprehensibility related to target emotion and steering strength (\(\lambda\)). 

To determine the inter-rater reliability among human evaluators, intraclass correlation coefficient (ICC) estimates were calculated based on single rater as well as mean of ten raters. Values between 0.5 and 0.75 were regarded as evidence of moderate reliability, values between 0.75 and 0.90 of good reliability, and values exceeding 0.90 of excellent reliability as suggested by~\cite{koo2016icc}.

\begin{figure*}[htbp]

\centering
\subfloat[Anger (human)]{\includegraphics[width=0.3\textwidth]{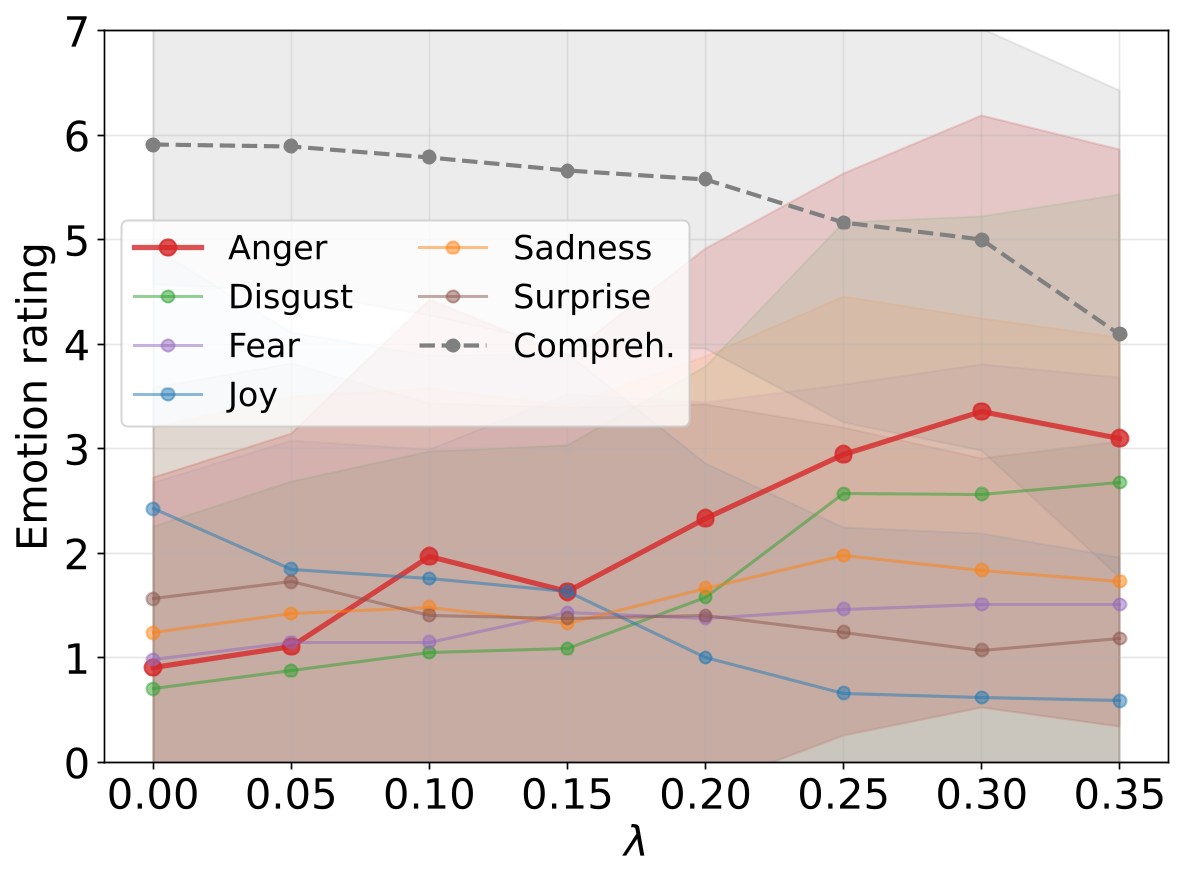}}
\hfill
\subfloat[Disgust (human)]{\includegraphics[width=0.3\textwidth]{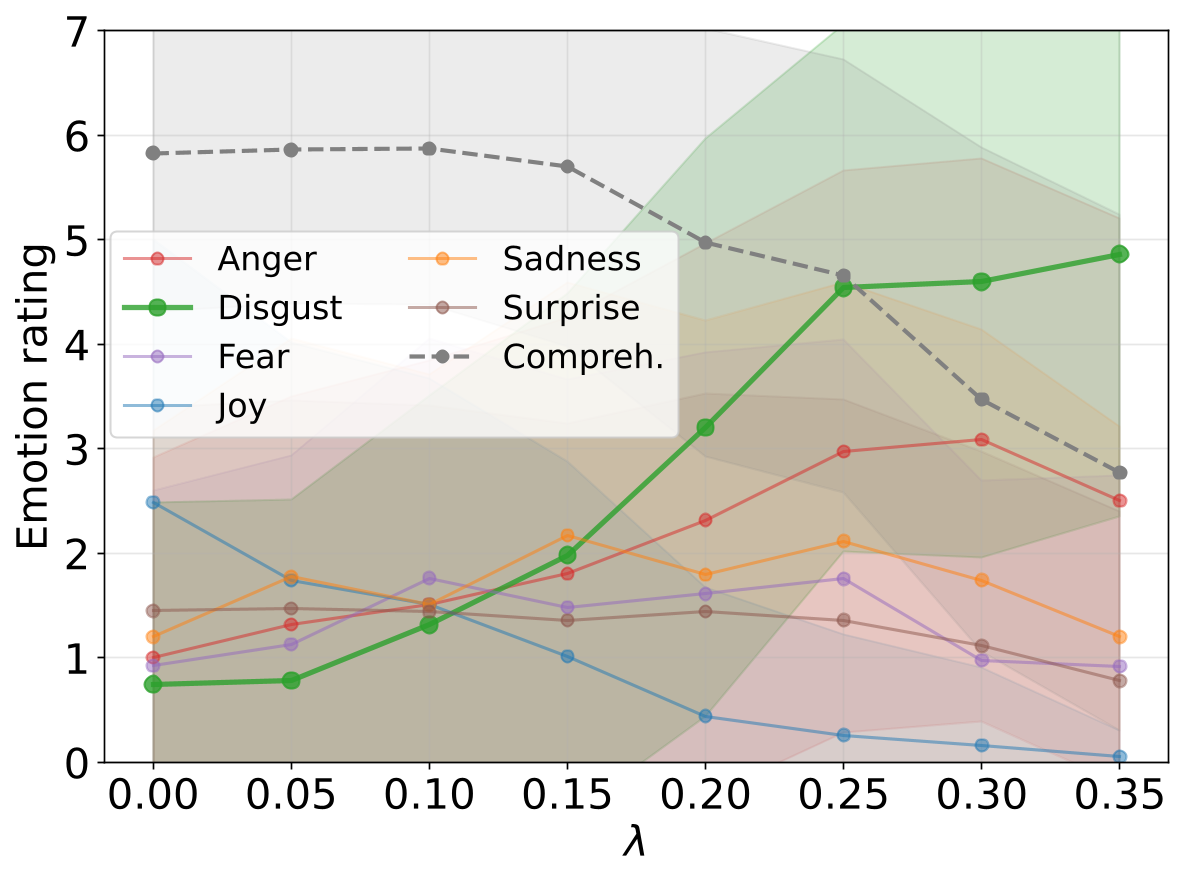}}
\hfill
\subfloat[Fear (human)]{\includegraphics[width=0.3\textwidth]{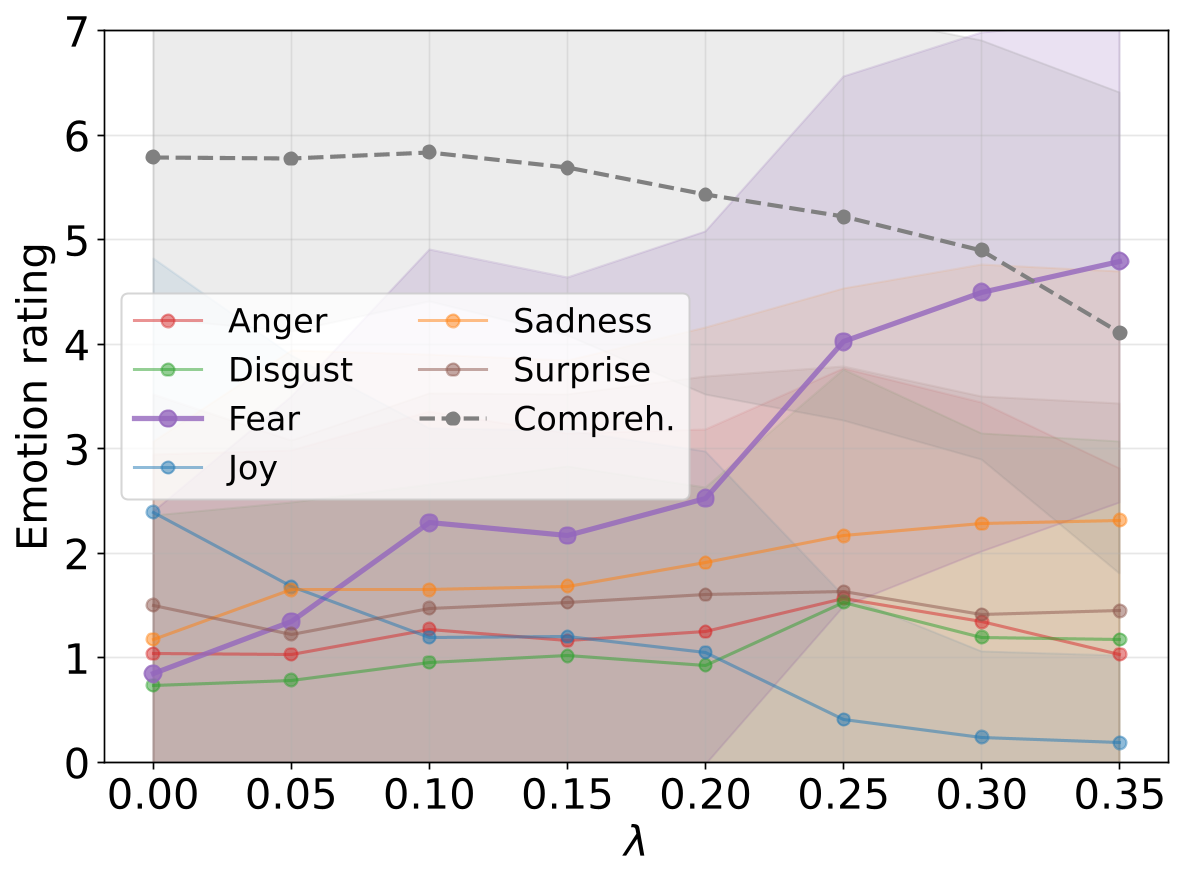}}

\vskip0.001\baselineskip

\subfloat[Anger (model)]{\includegraphics[width=0.3\textwidth]{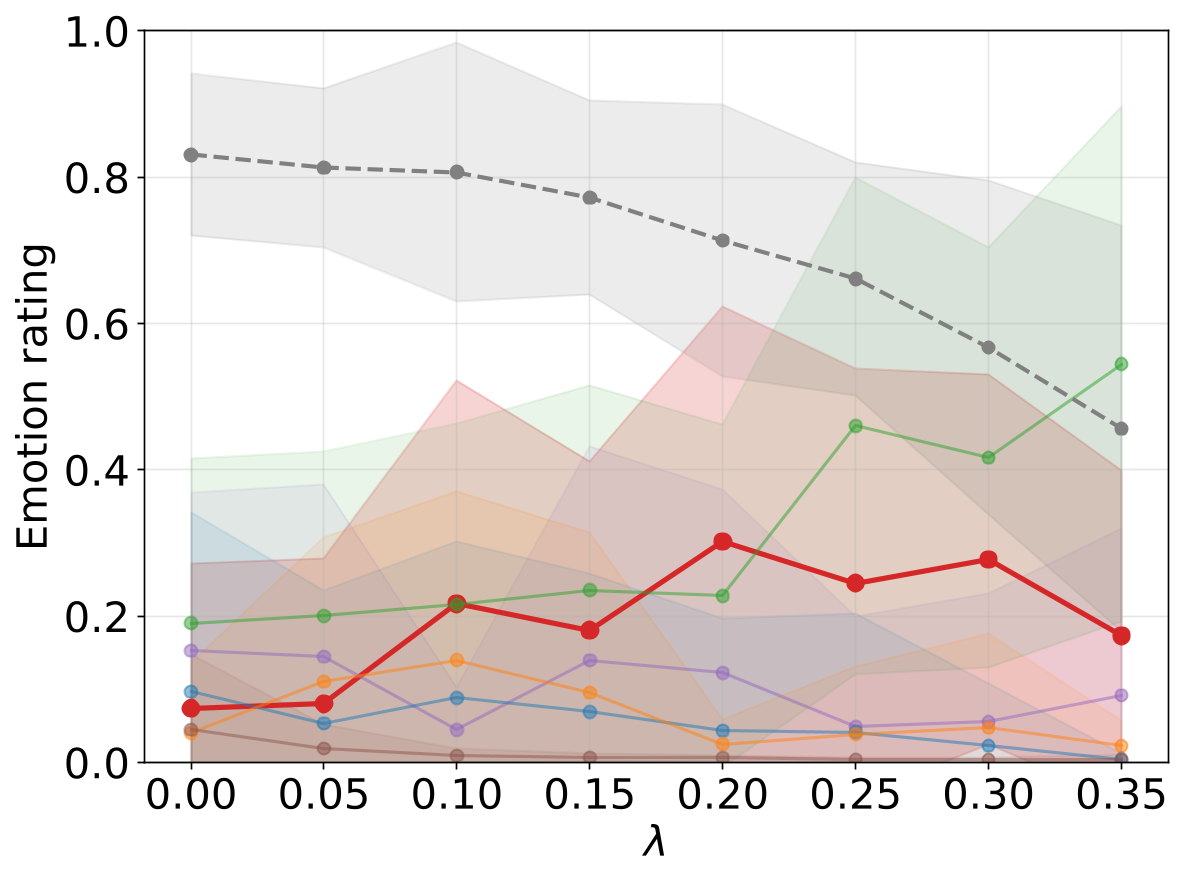}}
\hfill
\subfloat[Disgust (model)]{\includegraphics[width=0.3\textwidth]{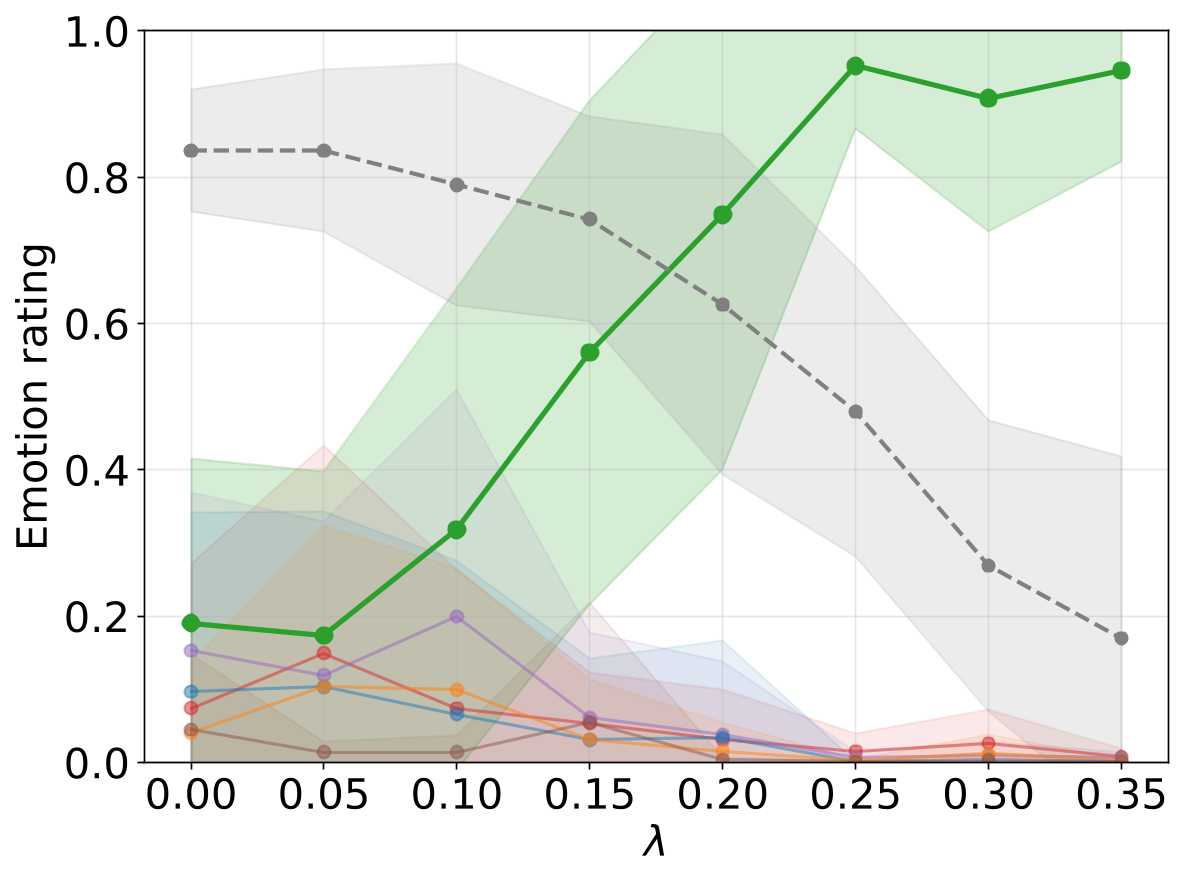}}
\hfill
\subfloat[Fear (model)]{\includegraphics[width=0.3\textwidth]{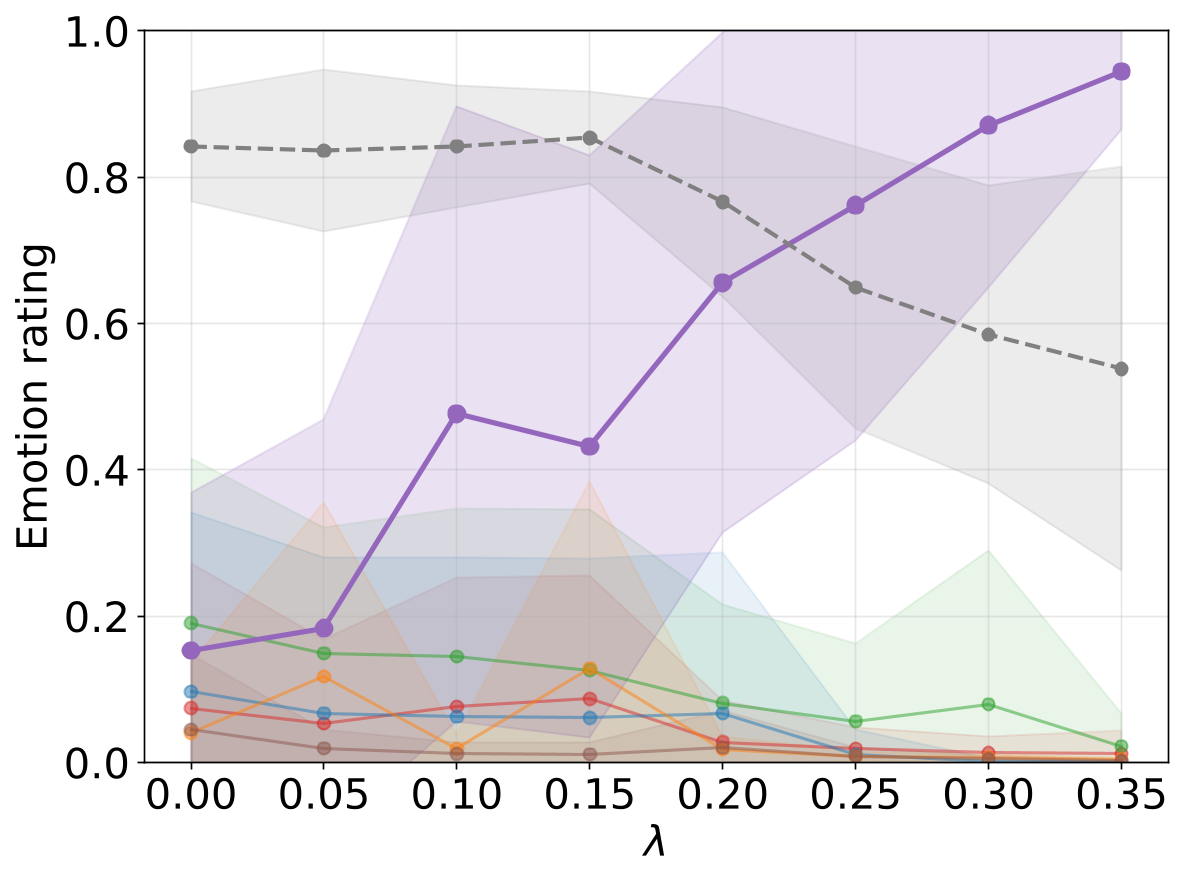}}

\vskip0.5\baselineskip
\centerline{\hdashrule[0.5ex]{0.9\linewidth}{0.5pt}{3pt 2pt}} 

\subfloat[Joy (human)]{\includegraphics[width=0.3\textwidth]{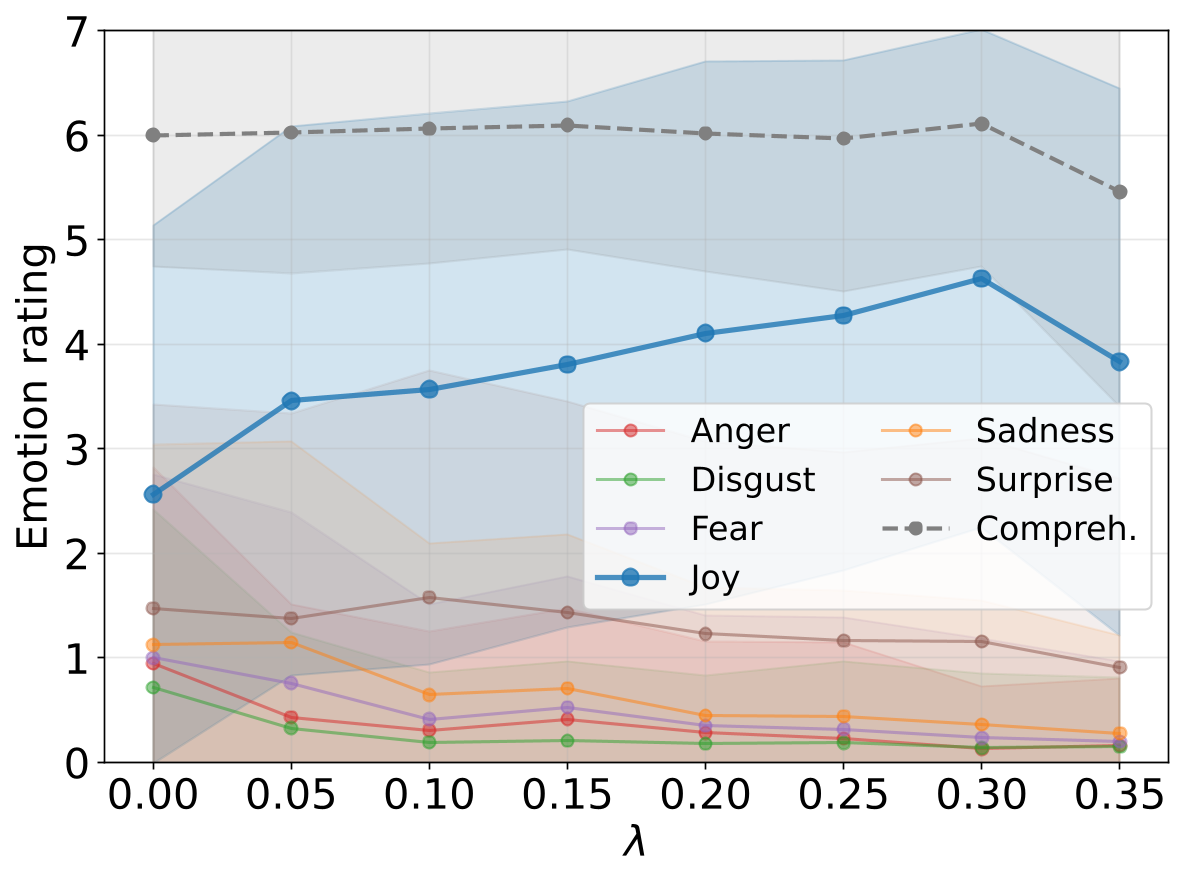}}
\hfill
\subfloat[Sadness (human)]{\includegraphics[width=0.3\textwidth]{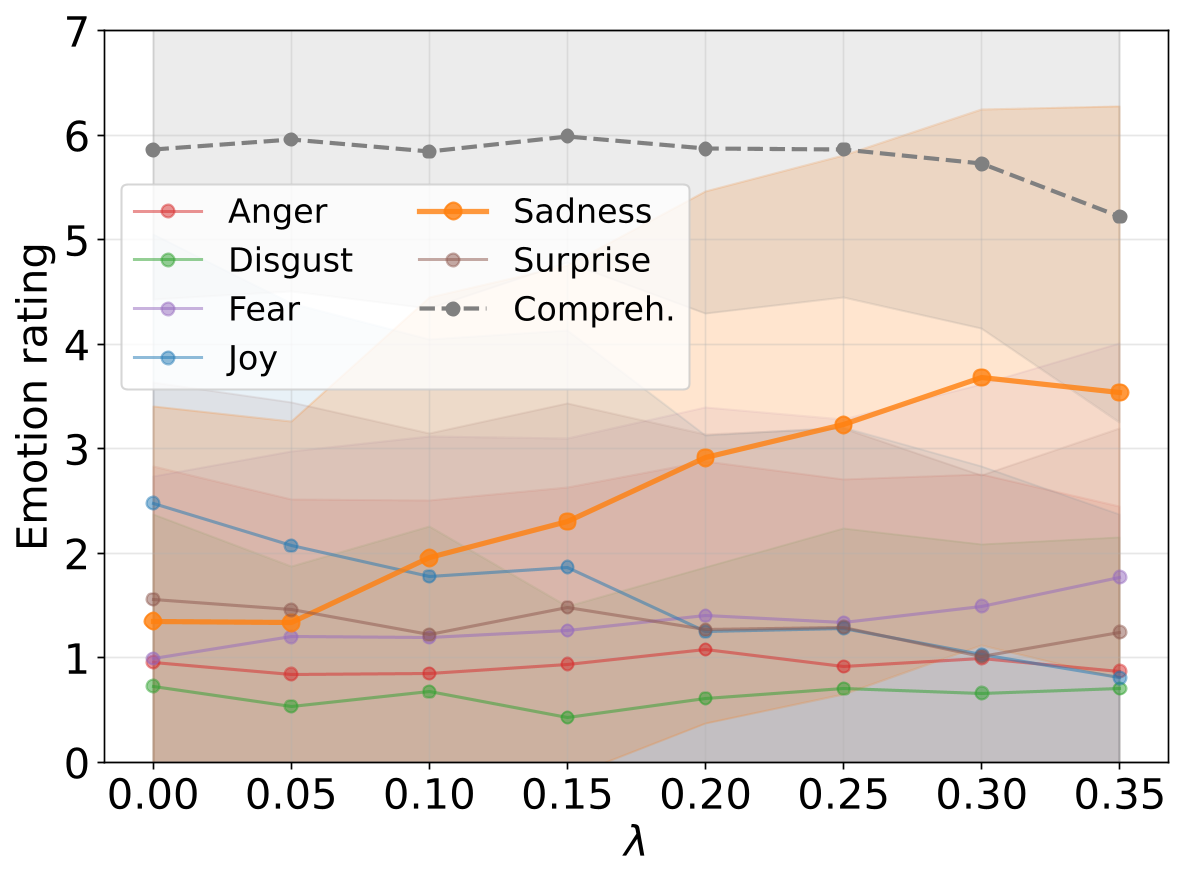}}
\hfill
\subfloat[Surprise (human)]{\includegraphics[width=0.3\textwidth]{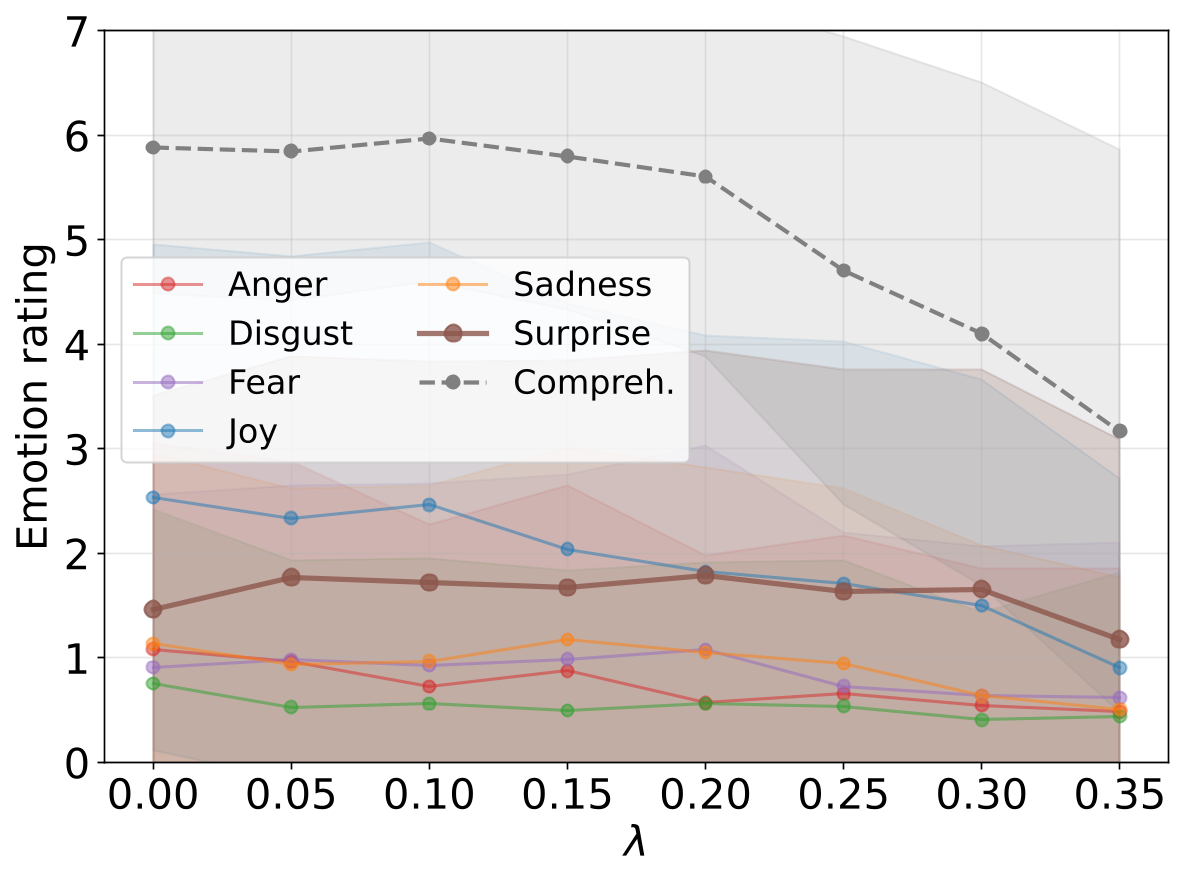}}

\vskip0.01\baselineskip

\subfloat[Joy (model)]{\includegraphics[width=0.3\textwidth]{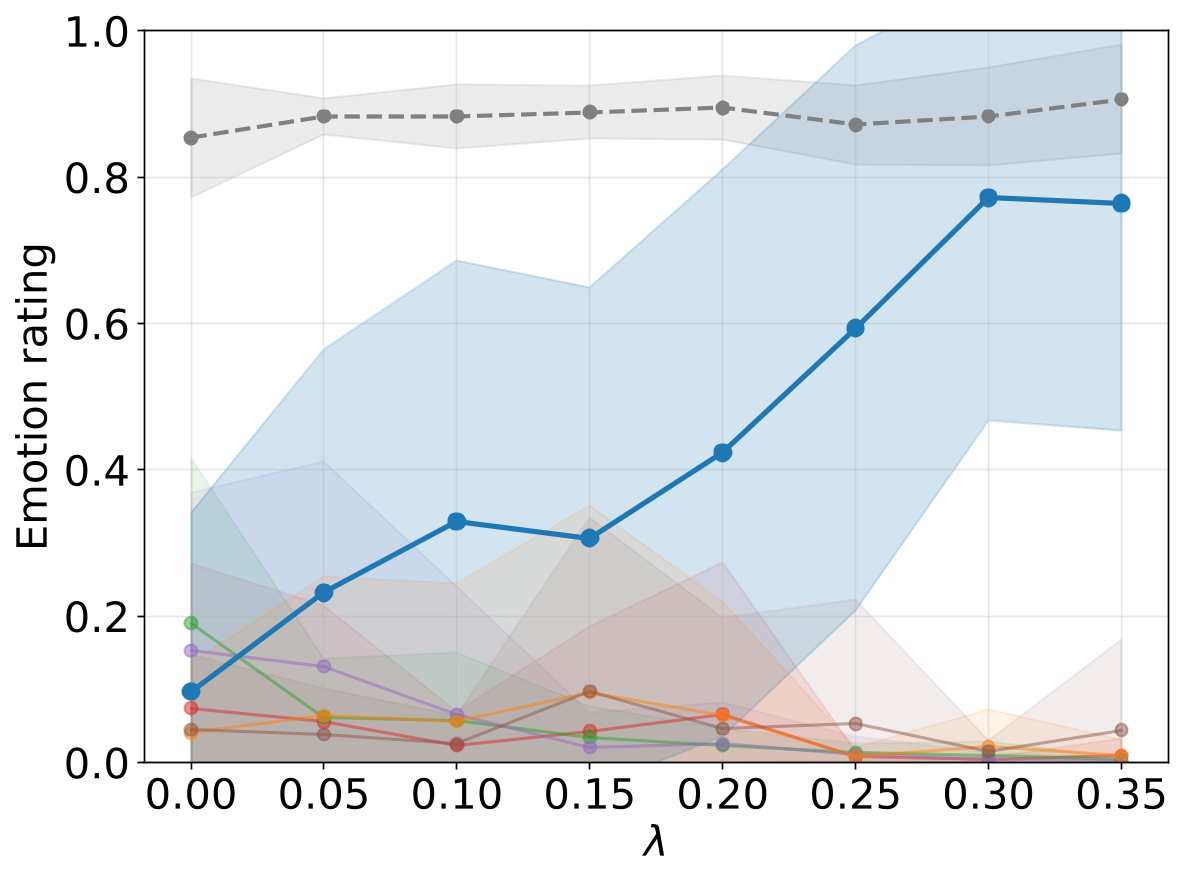}}
\hfill
\subfloat[Sadness (model)]{\includegraphics[width=0.3\textwidth]{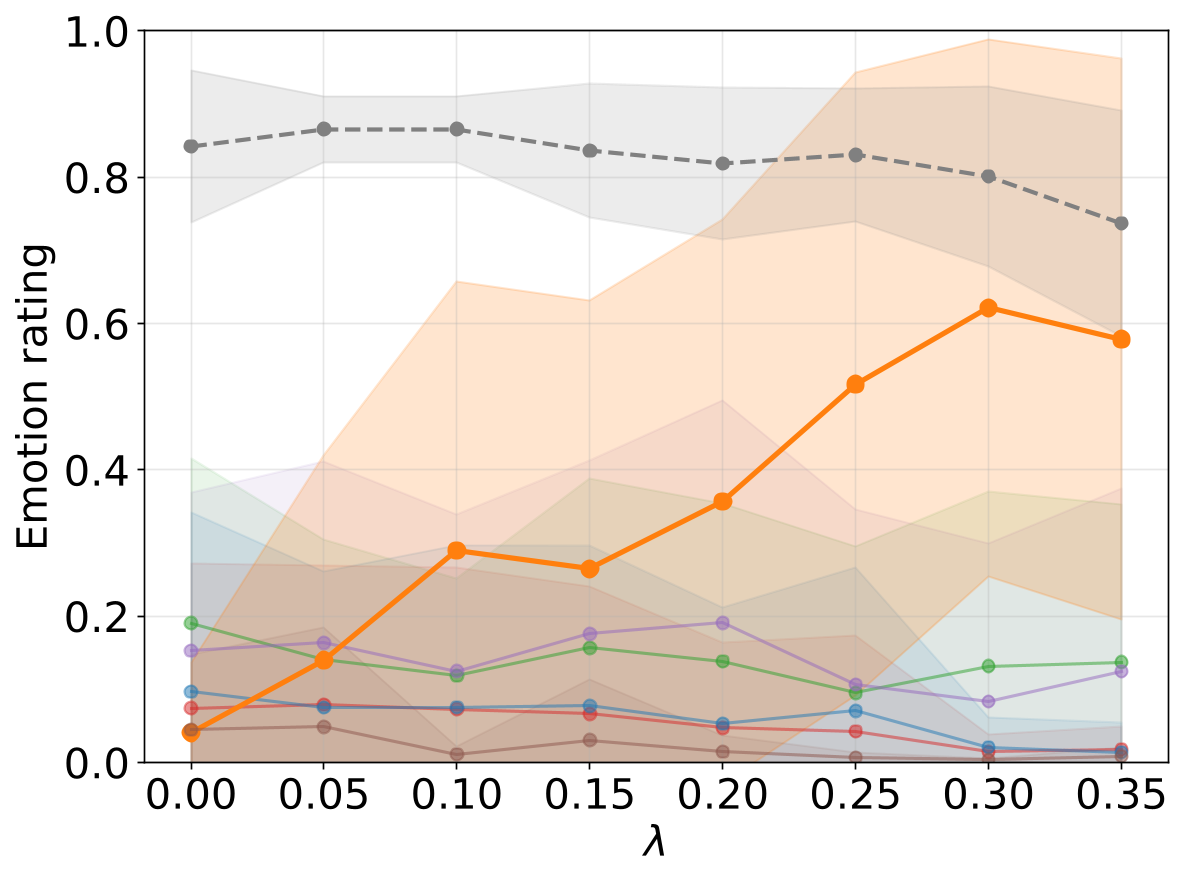}}
\hfill
\subfloat[Surprise (model)]{\includegraphics[width=0.3\textwidth]{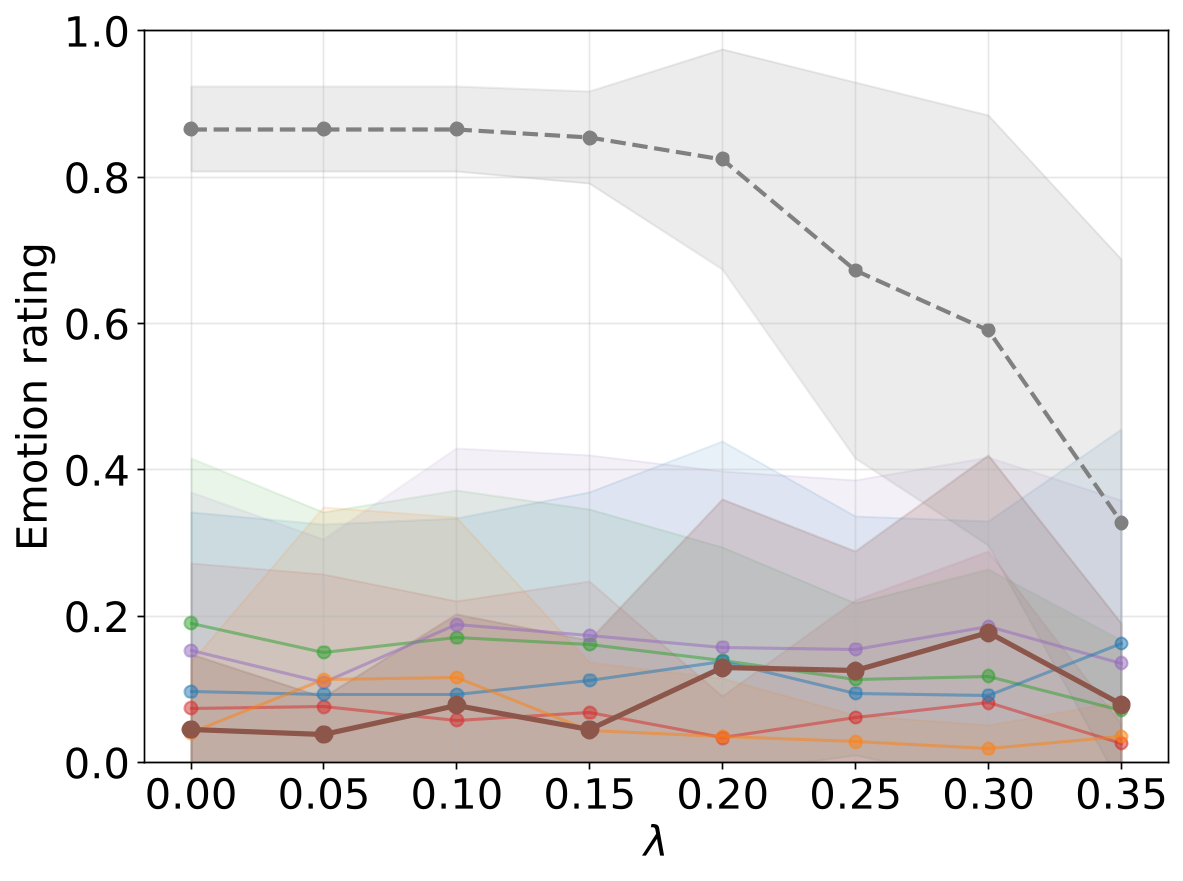}}

\caption{Human and model emotional perception and text comprehensibility ratings across steering strengths \(\lambda\) for six emotions. In each block, the top row shows average human mean ratings and the row below shows corresponding model scores. Shaded areas represent $\pm$1 standard deviation.}

\label{fig:emotion_steering_ratings}
\end{figure*}

\section{Results of Emotion Steering}
\label{sec:results_steering}

To assess the effect of varying steering strength on perceived emotion in generated text, we averaged emotion ratings over all participants and prompts for each \(\lambda\)-value and target emotion. In all experiments, steering vectors were applied across all layers. We used a fixed set of 19 prompts to elicit emotionally diverse responses, e.g., “Comment on a review of a business written by a customer”. The complete list of prompts is provided in Appendix~\ref{app:prompts}. Prior work and our preliminary experiments show that the effect of \(\lambda\) strongly depends on the number of layers to which steering is applied. All results and \(\lambda\)-based interpretations in this paper refer to the all-layer steering configuration.

Fig.~\ref{fig:emotion_steering_ratings} displays the mean human ratings alongside our automatic emotion classification model outputs. While participants provided ratings on a scale from 0 (no emotion) to 7 (maximum intensity), the model scores range from 0 to 1. Each line plot displays the ratings for one of six Ekman emotions, with the target emotion highlighted in bold and non-target emotions shown with lower opacity. Shaded areas in the plots represent $\pm$1 standard deviation across participants or model predictions, respectively. The figure also includes comprehensibility ratings. Participants rated the comprehensibility of each text on the same 0--7 scale (higher is better), while the model's comprehensibility predictor outputs scores from 1 (high comprehensibility) to 10 (low). We normalized and inverted these model comprehensibility scores for visual alignment.

\subsection{Emotional Intensity Across Steering Strengths}

Overall, the human ratings revealed a clear and consistent increase in perceived emotional intensity, with higher \(\lambda\)-values for five of the six emotions. The effect was particularly pronounced for \emph{disgust} and \emph{fear}, both of which rose from an average rating of 0.75 and 0.85 (at \(\lambda = 0.00\)) to 4.86 and 4.80 (at \(\lambda = 0.35\)). \emph{Joy} also showed a positive trend, increasing from 2.57 to 3.83 with a peak at \(\lambda = 0.30\) reaching 4.63. \emph{Anger} and \emph{sadness} followed similar, though more moderate, trends—anger increased from 0.91 to 3.10, and \textit{sadness} from 1.35 to 3.54. Notably, \emph{surprise} did not exhibit a meaningful change in perceived intensity (from 1.47 to 1.18), suggesting that it is less effectively steered via our method. Mean ratings are also summarized in Table~\ref{tab:descriptive_stats} in Appendix~\ref{app:descriptive_stats}.

To test our observations statistically, we conducted a two-way repeated-measures ANOVA. Mauchly's test indicated that the assumption of sphericity was violated for all effects and all but two \(\epsilon\) values were $<$ 0.75, Greenhouse-Geisser correction was applied to the degrees of freedom for all effects. Results of the ANOVAs (Table~\ref{tab:anova}) revealed significant main effects of target emotion (all \textit{p}s $<$ 0.001), steering strength (all \textit{p}s $<$ 0.001) as well as significant interactions between target emotion and steering strength, suggesting that the effect of \(\lambda\) depends, to some extent, on emotion. The effect of emotion was overall stronger than the effect of steering strength and strongest in perceived emotional intensity of \textit{joy} ($\eta_p^2$ = 0.618) and \textit{disgust} ($\eta_p^2$ = 0.616) and weakest in perceived emotional intensity of \textit{surprise} ($\eta_p^2$ = 0.042). In line with our assumptions, we found a relatively strong effect of steering strength on text comprehensibility ($\eta_p^2$ = 0.488). Overall, these results support the effectiveness of emotion steering for five of the six emotions.

\begin{table}
    \centering
    \caption{Two-way repeated-measures Analysis of Variance (ANOVA) in perceived emotional intensity and text comprehensibility}
    \begin{threeparttable}
        \begin{tabular}{l r r r r r}
        \toprule
                                        & df1       & df2       & \textit{F}& \textit{p}& $\eta_p^2$ \\
        \midrule
        Anger                           &           &           &           &           &  \\
        \midrule
        Target emotion (E)              &  3.12     &  589.81   & 161.61    & $<$.001   &  .461 \\
        Steering strength (\(\lambda\)) &  3.50     &  661.71   &  20.29    & $<$.001   &  .097 \\
        E x \(\lambda\)                 & 19.67     & 3716.64   &  24.93    & $<$.001   &  .117 \\
        \midrule
        Disgust                         &           &           &           &           &  \\
        \midrule
        Target emotion (E)              &  3.00     &  567.40   & 303.21    & $<$.001   & .616 \\
        Steering strength (\(\lambda\)) &  4.07     &  768.53   &  77.55    & $<$.001   & .291 \\
        E x \(\lambda\)                 & 18.19     & 3437.59   &  58.93    & $<$.001   & .238 \\
        \midrule
        Fear                            &           &           &           &           &  \\
        \midrule
        Target emotion (E)              &  3.53     &  667.54   & 222.26    & $<$.001   & .540 \\
        Steering strength (\(\lambda\)) &  4.84     &  914.82   &  24.16    & $<$.001   & .113 \\
        E x \(\lambda\)                 & 21.97     & 4151.99   &  41.15    & $<$.001   & .179 \\
        \midrule
        Joy                             &           &           &           &           &  \\
        \midrule
        Target emotion (E)              &  2.56     &  484.35   & 305.91    & $<$.001   & .618 \\
        Steering strength (\(\lambda\)) &  2.98     &  562.98   &  54.37    & $<$.001   & .223 \\
        E x \(\lambda\)                 & 22.72     & 4293.67   &  26.98    & $<$.001   & .125 \\
        \midrule
        Sadness                         &           &           &           &           &  \\
        \midrule
        Target emotion (E)              &  3.52     &  664.90   & 124.08    & $<$.001   & .396 \\
        Steering strength (\(\lambda\)) &  4.55     &  859.59   &  13.87    & $<$.001   & .068 \\
        E x \(\lambda\)                 & 22.69     & 4288.75   &  20.43    & $<$.001   & .098 \\
        \midrule
        Surprise                        &           &           &           &           &  \\
        \midrule
        Target emotion (E)              &  4.08     &  771.63   &   8.19    & $<$.001   & .042 \\
        Steering strength (\(\lambda\)) &  4.90     &  926.85   &   8.56    & $<$.001   & .043 \\
        E x \(\lambda\)                 & 24.66     & 4659.96   &   2.55    & $<$.001   & .013 \\
        \midrule
        Comprehensibility               &           &           &           &           &  \\
        \midrule
        Target emotion (E)              &  4.20     &  794.17   & 100.20    & $<$.001   & .346 \\
        Steering strength (\(\lambda\)) &  3.38     &  638.71   & 179.85    & $<$.001   & .488 \\
        E x \(\lambda\)                 & 20.31     & 3838.68   &  22.78    & $<$.001   & .108 \\ 
        \bottomrule 
        \end{tabular}
        \begin{tablenotes}
            \item Greenhouse-Geisser corrected degrees of freedom for all effects.
        \end{tablenotes}
    \end{threeparttable}
    \label{tab:anova}
\end{table}

The automatic model ratings confirmed and even amplified the trends observed in our sample of human evaluators. \emph{Disgust}, \emph{fear}, and \emph{joy} rose sharply from approximately 0.1–0.2 at \(\lambda = 0.00\) to scores near 0.9–1.0 at \(\lambda = 0.35\). \emph{Sadness} showed a moderate increase up to about 0.6. Notably, \emph{anger} appeared to be less effectively captured by the classifier, showing only a minor increase from around 0.05 to 0.2, peaking early at \(\lambda = 0.20\). Instead, the steering of anger seemed to clearly affect the disgust emotion score, starting at 0.2 and reaching a score slightly above 0.5. \emph{Surprise}, in line with the human ratings, displayed no or only a minimal trend across the steering range.

Interestingly, a key difference emerged when comparing human and model ratings for non-target emotions. In the case of the automatic emotion model, ratings for non-target emotions remained largely stable or even decreased as the steering strength (\(\lambda\)) increased. This suggests that the model is highly selective in interpreting emotional expression, predominantly increasing only the intended emotion dimension while suppressing or ignoring unrelated ones.

By contrast, human ratings revealed a more complex and overlapping emotion perception. For instance, when steering towards \emph{anger}, participants not only rated \emph{anger} higher but also reported increased levels of perceived \emph{disgust} and \emph{sadness}, with a minor increase in \emph{fear}. Similarly, steering towards \emph{disgust} caused higher ratings for both \emph{anger} and—less consistently—\emph{sadness} and \emph{fear}. In the case of \emph{fear}, participants also perceived elevated levels of \emph{sadness}, further highlighting the interconnected nature of specific emotional categories.

These findings reflect the inherently multidimensional and context-dependent nature of human emotion perception. Unlike the model’s interpretation, human raters tend to perceive emotional states as overlapping or co-occurring, especially among low-valence emotional expressions ~\cite{russell1980,watson1988,larsen2001,barrett2004,scherer2009}. The results emphasize that while activation-based steering successfully amplifies target emotions, the resulting texts often carry nuanced emotional undertones that human perception picks up, even when these are not explicitly modeled.

\begin{figure}[t]
  \centering
  \includegraphics[width=0.9\linewidth]{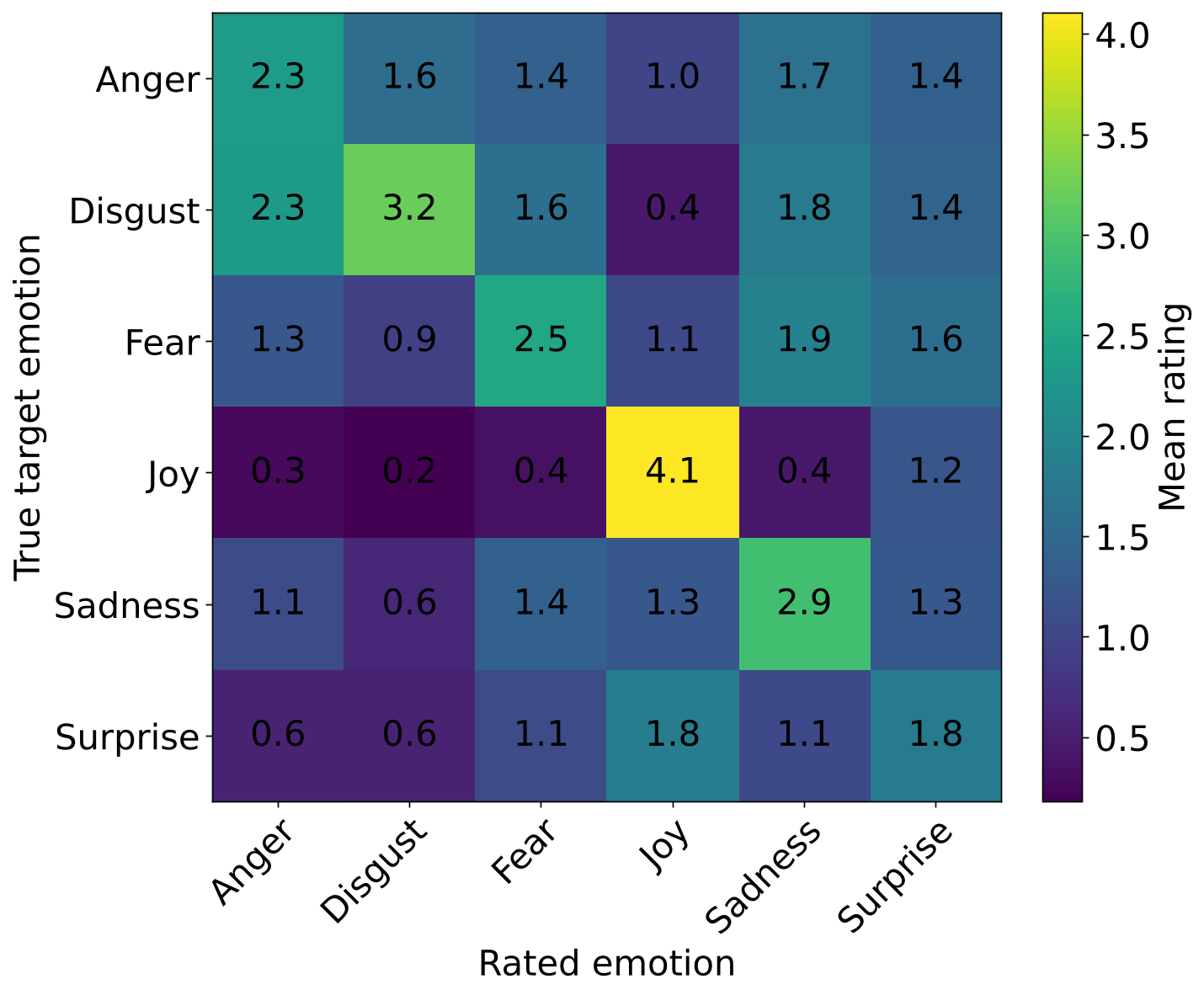}
  \caption{Participants’ mean ratings heatmap at $\lambda=0.20$. Rows denote the true target emotion; columns denote the average rating participants assigned to each emotion dimension (0--7 scale).}
  \label{fig:participants_mean_heatmap}
\end{figure}

To quantify co-perception of emotions, Fig.~\ref{fig:participants_mean_heatmap} shows a heatmap of mean human ratings at moderate steering strength $\lambda = 0.20$. For each target emotion (rows), the six rated intensities (0--7) across emotion dimensions (columns) were averaged. Strong diagonals indicate successful steering, especially for \emph{joy}. Off-diagonal clusters reveal systematic co-activation among low-valence emotions: \emph{anger} aligns with \emph{disgust}, \emph{sadness}, and \emph{fear}, and \emph{fear} aligns with \emph{sadness}. \emph{Surprise} remains weak and diffuse. These patterns mirror the trajectories in Figs.~\ref{fig:emotion_steering_ratings}: steering a target emotion increases its perceived intensity while also elevating related low-valence dimensions, highlighting the interconnected perception of affective categories.

Despite the consistent upward trends across participants, the shaded areas in the plots indicate substantial variance in individual emotion ratings. This highlights that while the average perception of emotional intensity increases with higher \(\lambda\), participants differed in how strongly they perceived the emotional tone of the generated texts. These differences may stem not only from personal interpretation styles, cultural or linguistic nuances, or varying thresholds for recognizing emotion in written language, but also from differences in the prompts themselves. Thus, the results should reflect population-level trends rather than uniform perceptions across all raters.

\begin{figure*}[htbp]
\centering

\subfloat[Anger]{\includegraphics[width=0.3\textwidth]{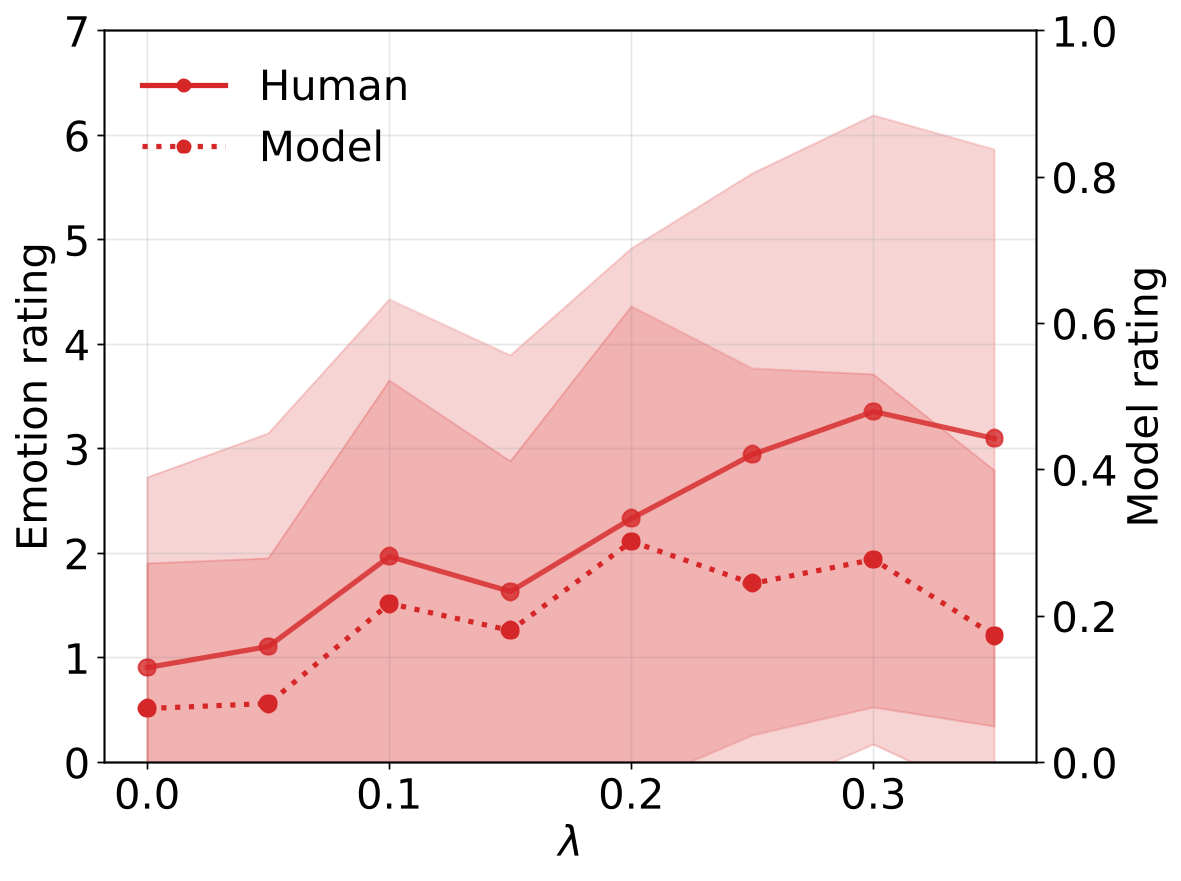}}
\hfill
\subfloat[Disgust]{\includegraphics[width=0.3\textwidth]{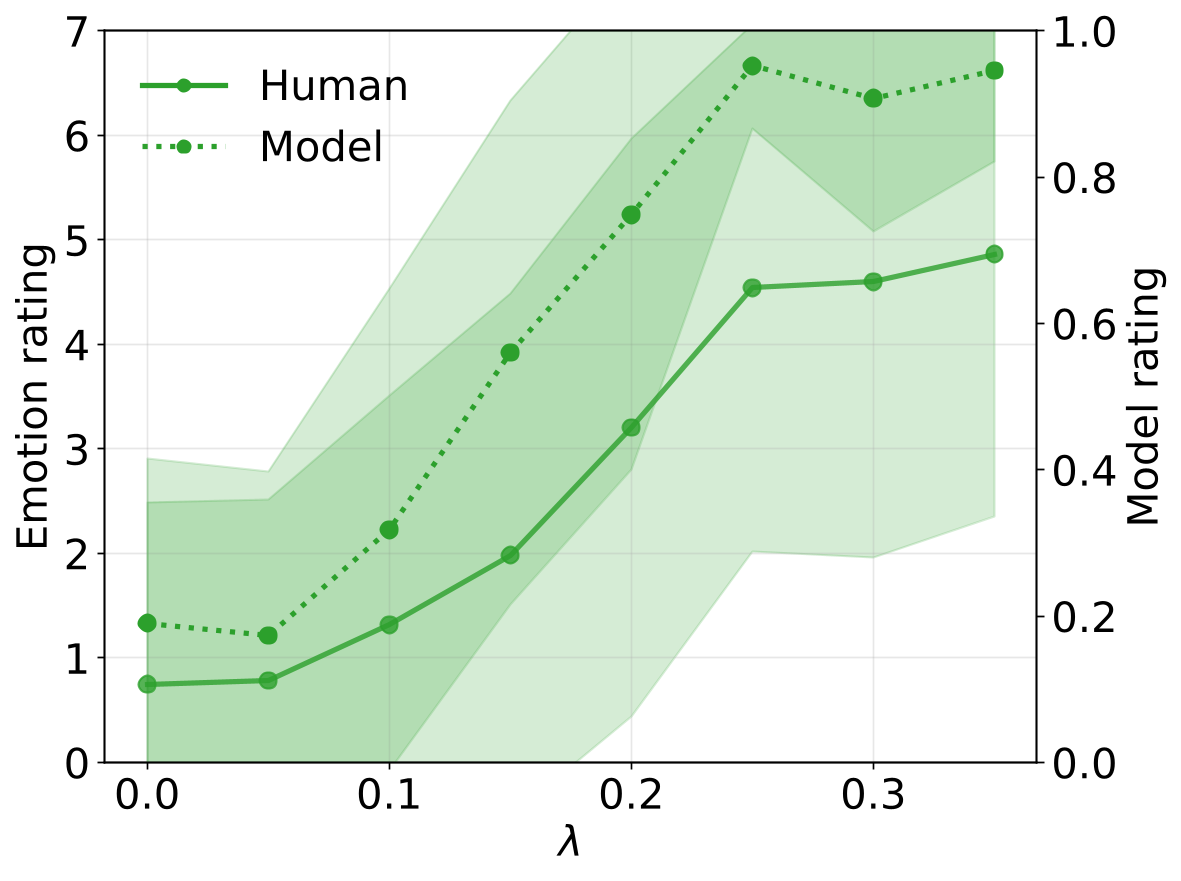}}
\hfill
\subfloat[Fear]{\includegraphics[width=0.3\textwidth]{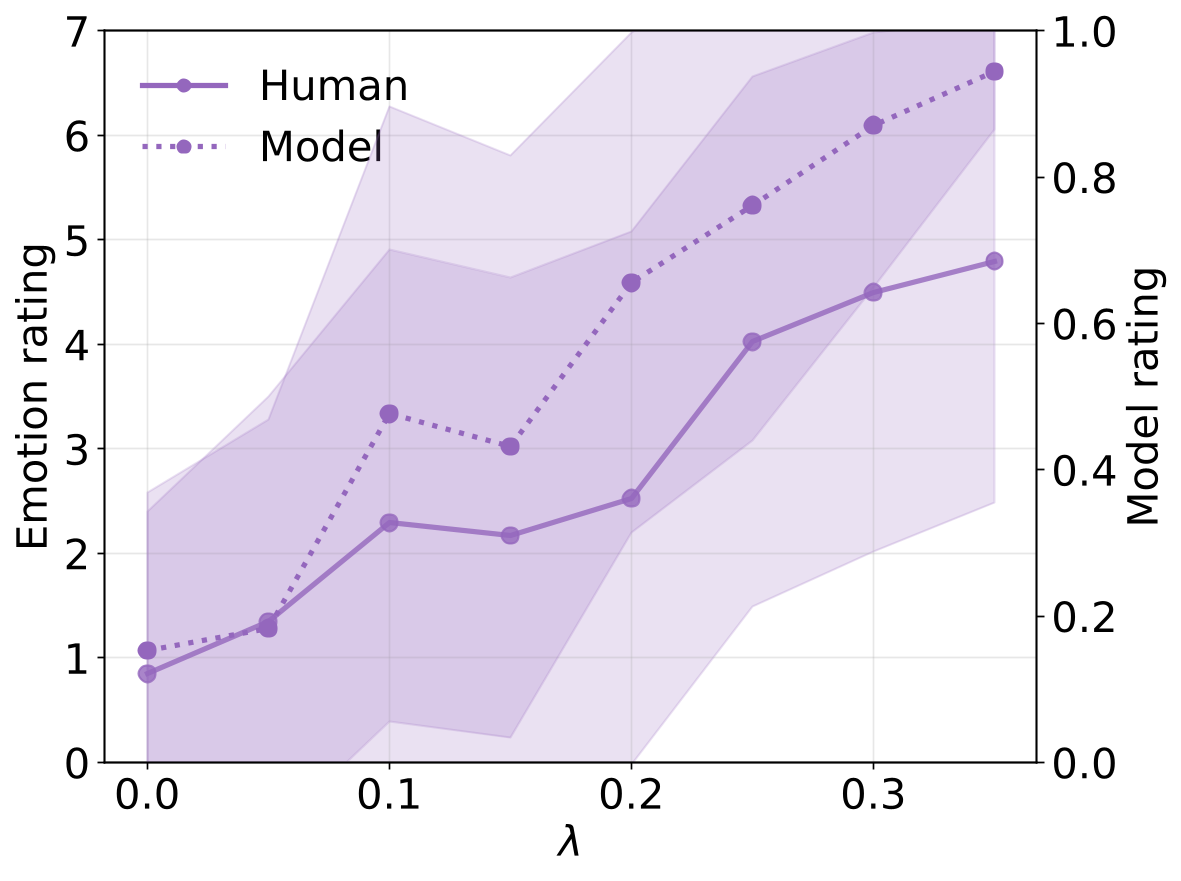}}

\subfloat[Joy]{\includegraphics[width=0.3\textwidth]{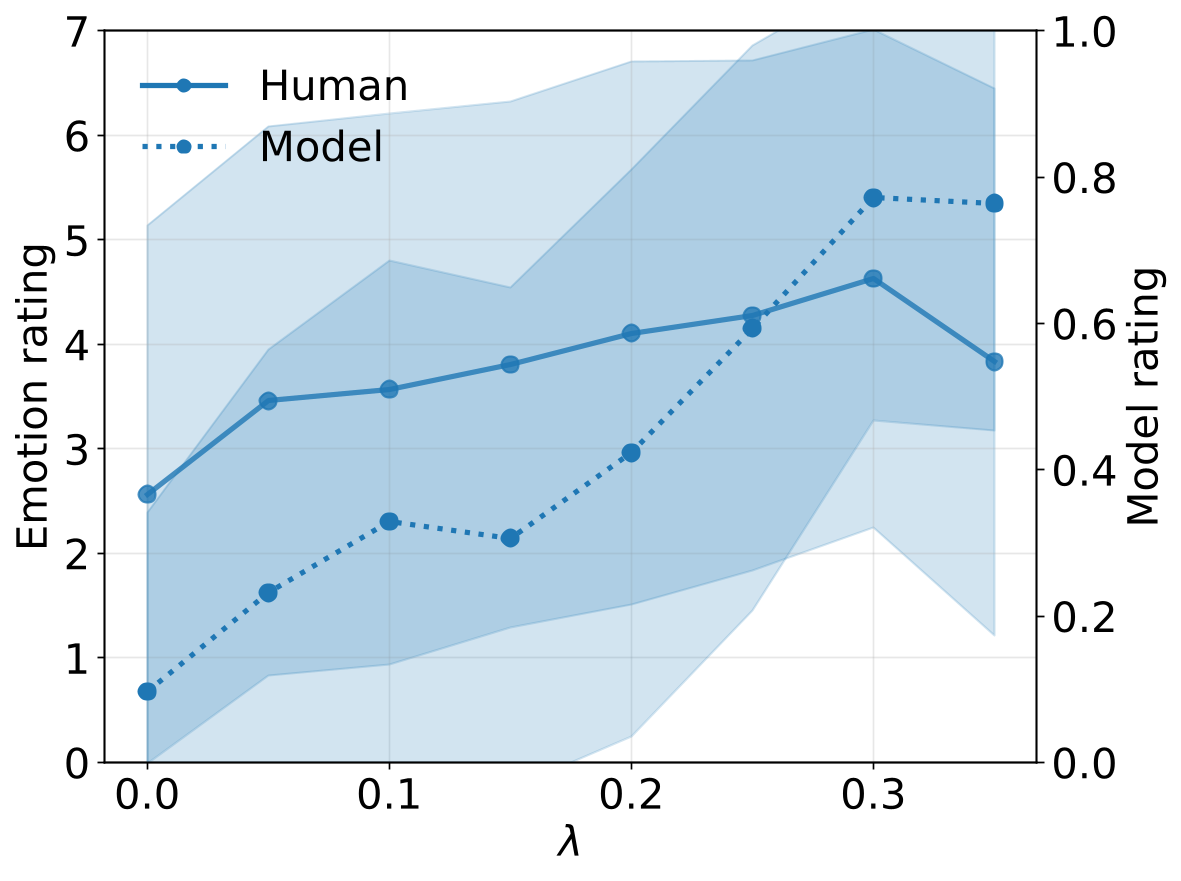}}
\hfill
\subfloat[Sadness]{\includegraphics[width=0.3\textwidth]{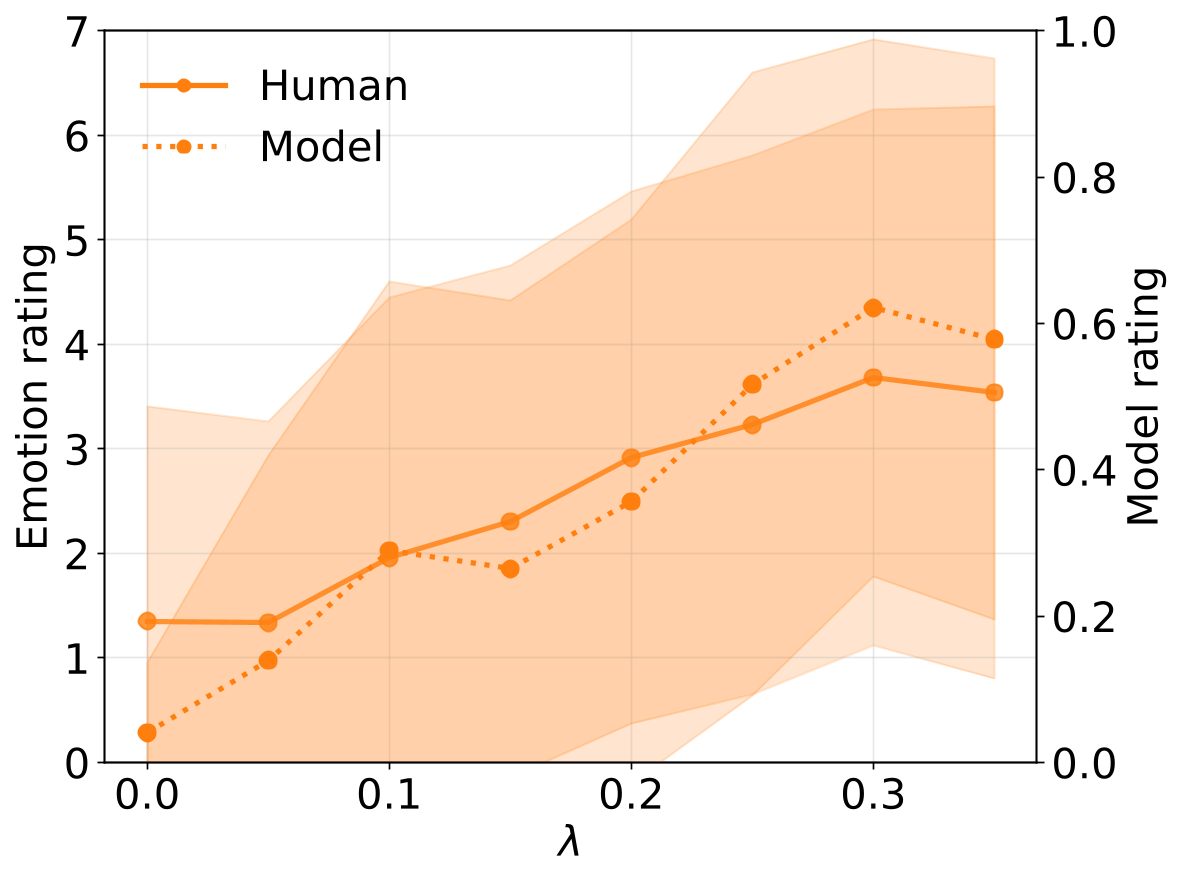}}
\hfill
\subfloat[Surprise]{\includegraphics[width=0.3\textwidth]{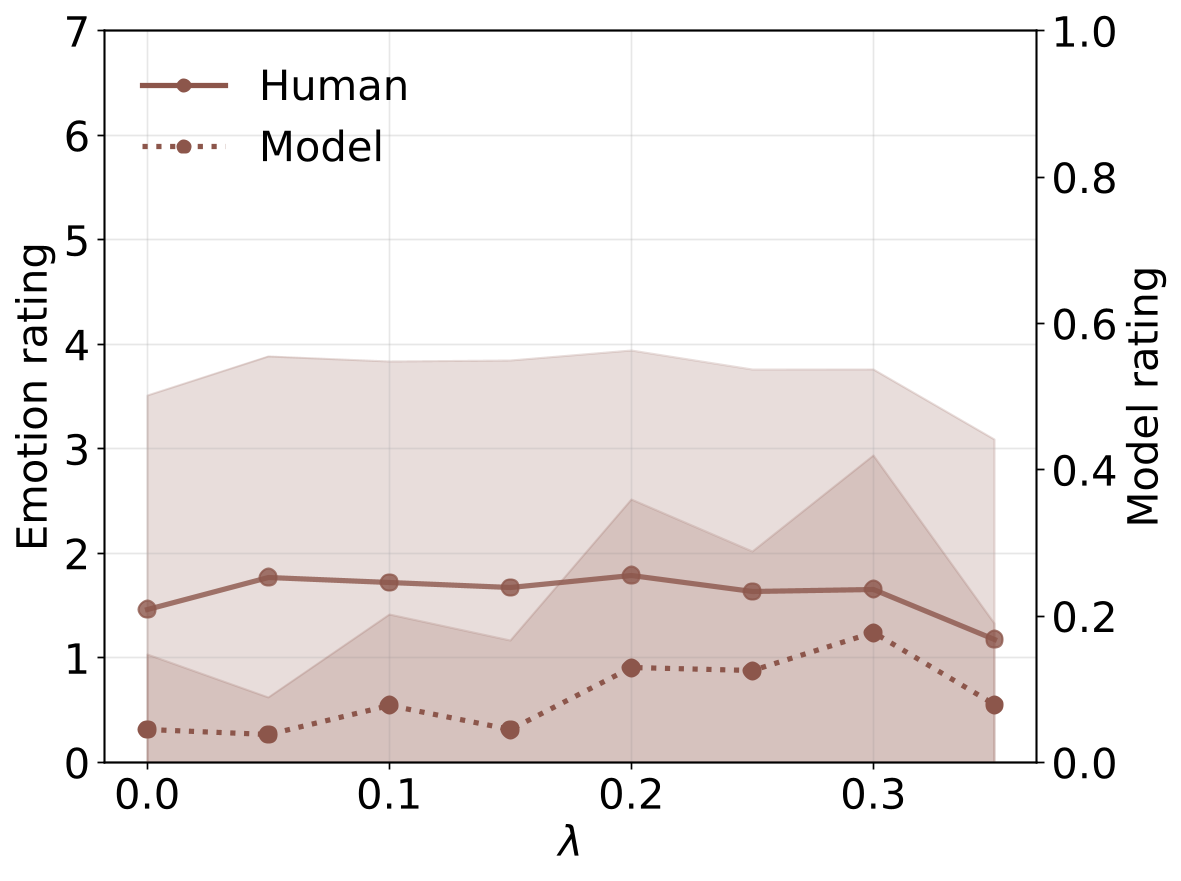}}

\caption{Comparison of human emotion perception and model emotion steering ratings across six target emotions. Each plot shows the average perceived intensity of the target emotion as a function of steering strength (\(\lambda\)). Human ratings (left y-axis, 0--7 scale) are plotted alongside model predictions (right y-axis, 0--1 scale). Shaded areas represent $\pm$1 standard deviation.}
\label{fig:emotion_target_comparison}
\end{figure*}

To determine the consistency of human ratings, we quantified inter-rater reliability using intraclass correlations (ICCs), reporting both single-rater \mbox{ICC(2,1)} and aggregated-rater \mbox{ICC(2,$k$)} values. ICCs were computed within each prompt × target-emotion block and averaged across prompts. As summarized in Table~\ref{tab:icc}, single-rater reliability is modest (\textit{M}\,$\approx$\,0.24–0.43 across five emotions), while aggregated reliability is high for \emph{anger}, \emph{disgust}, \emph{fear}, \emph{joy}, and \emph{sadness} (\textit{M}\,$\approx$\,0.71–0.87). In contrast, \emph{surprise} shows low agreement ($\approx$ 0.28), reinforcing that this emotion is weakly and inconsistently perceived in our setting—consistent with its limited steerability observed in Figs.~\ref{fig:emotion_steering_ratings} and~\ref{fig:emotion_target_comparison}, and Table \ref{tab:anova}. These results suggest that although individual raters vary considerably in their ratings, the aggregated judgments reveal consistent population-level trends.

\begin{table}
\centering
\caption{\textbf{Inter-rater reliability} (means across 19 prompts per emotion). \mbox{ICC(2,1)}: single rater; \mbox{ICC(2,$k$)}: mean of $k{=}10$ raters.}
\label{tab:icc}
\setlength{\tabcolsep}{10pt}
\begin{tabular}{lcc}
\hline
\textbf{Emotion} & \textbf{ICC(2,1)} & \textbf{ICC(2,$k$)} \\
\hline
Anger    & 0.33 & 0.79 \\
Disgust  & 0.43 & 0.87 \\
Fear     & 0.36 & 0.81 \\
Joy      & 0.42 & 0.85 \\
Sadness  & 0.24 & 0.71 \\
Surprise & 0.05 & 0.28 \\
\hline
\end{tabular}
\end{table}

\subsection{Alignment of Human Judgments and Automatic Classifier}

To further examine the alignment between human perception and the model's emotion estimates, we directly compared average ratings for each target emotion across increasing \(\lambda\)-values, as shown in Fig.~\ref{fig:emotion_target_comparison}. Each plot contains one line for the human ratings (0--7 scale, left y-axis) and one for the model scores (0--1 scale, right y-axis).

The trends demonstrate a good alignment between model and human ratings as steering strength increases. Particularly for \emph{sadness}, \emph{disgust}, and \emph{fear}, the curves closely follow one another, indicating that the direction and shape of emotional amplification are comparable across both perspectives.

However, some divergences become apparent. For \emph{anger}, the deviation between model and human ratings began around \(\lambda = 0.2\), where human raters perceived less intensity than the model estimates. In the cases of \emph{disgust} and \emph{fear}, both ratings increased consistently with \(\lambda\), but the model scores rose more steeply, approaching the upper limit of 1, while human ratings remained lower. This suggests that humans apply more conservative upper-bound ratings while the emotional trajectory is shared.

To quantify the alignment between human and model ratings, we computed Pearson's correlation coefficient over each emotion's average ratings per \(\lambda\). The results revealed strong correlations for most emotions: \emph{disgust} (\textit{r} = .985), \emph{fear} (\textit{r} = .973), and \emph{sadness} (\textit{r} =.971) show high alignment. \emph{Joy} (\textit{r} = .813) and \emph{anger} (\textit{r} = .758) also demonstrate substantial correspondence, despite some scale mismatches. The only notable outlier was \emph{surprise}, with a much lower correlation of \textit{r} = .157, reinforcing the impression that this emotion is less consistently influenced by the steering mechanism. Across all six emotions, the average Pearson correlation was \textit{r} = .776, suggesting a generally strong relationship between human perception and model output.
In the case of \emph{joy}, an interesting pattern emerged: human raters already perceived a notable level of joy at \(\lambda = 0\) (around 2.5), while the model outputs remained near zero. As \(\lambda\) increases, the model's score rose more sharply than human perception, resulting in a crossover at approximately \(\lambda = 0.25\). A similar but less pronounced effect occurs for \emph{sadness}, where the model begins at a lower value but surpasses human ratings by the highest steering strength.
For \emph{surprise}, no meaningful change in perceived emotion occurred for humans or the model across steering levels, reinforcing previous findings that this emotion is less susceptible to our current steering method.
Across the board, the progression of ratings for \(\lambda\) tends to show similar acceleration in both human and model responses, suggesting that \(\lambda\) provides a reasonably consistent control signal. This consistency across emotion types reinforces the viability of using \(\lambda\) as a global steering parameter, even if perceived intensity ceilings differ between raters and automated classifiers.
Among the five emotions for which we observe an apparent steering effect, four exhibited a stronger final intensity in the model ratings than the human assessments (all except \emph{anger}). This indicates that human raters generally avoid extreme values in the upper range of the scale: in no case does the average human rating exceed 5, even at the highest steering level. The model, by contrast, tends to saturate more aggressively, particularly for emotions such as \emph{disgust} and \emph{fear}.

\begin{figure*}[htbp]
\centering

\subfloat[Anger]{\includegraphics[width=0.3\textwidth]{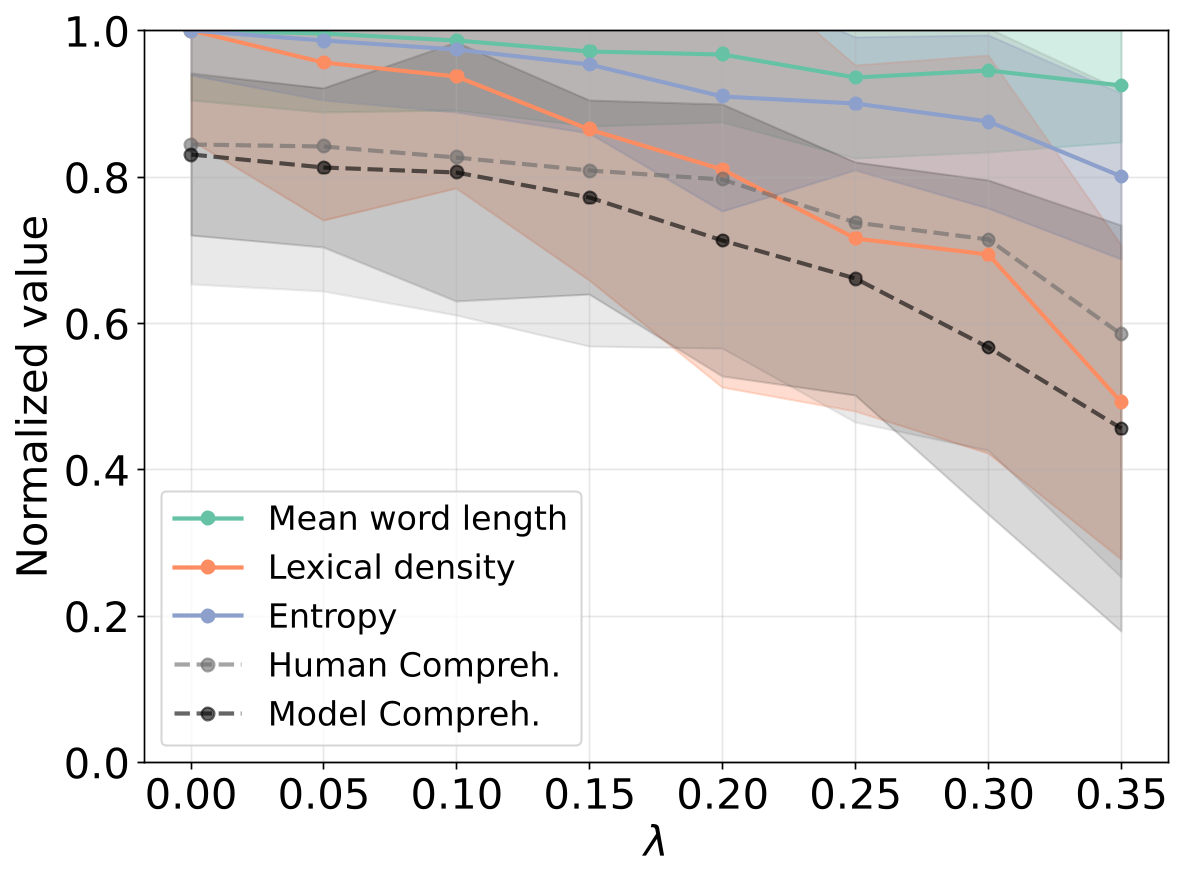}}
\hfill
\subfloat[Disgust]{\includegraphics[width=0.3\textwidth]{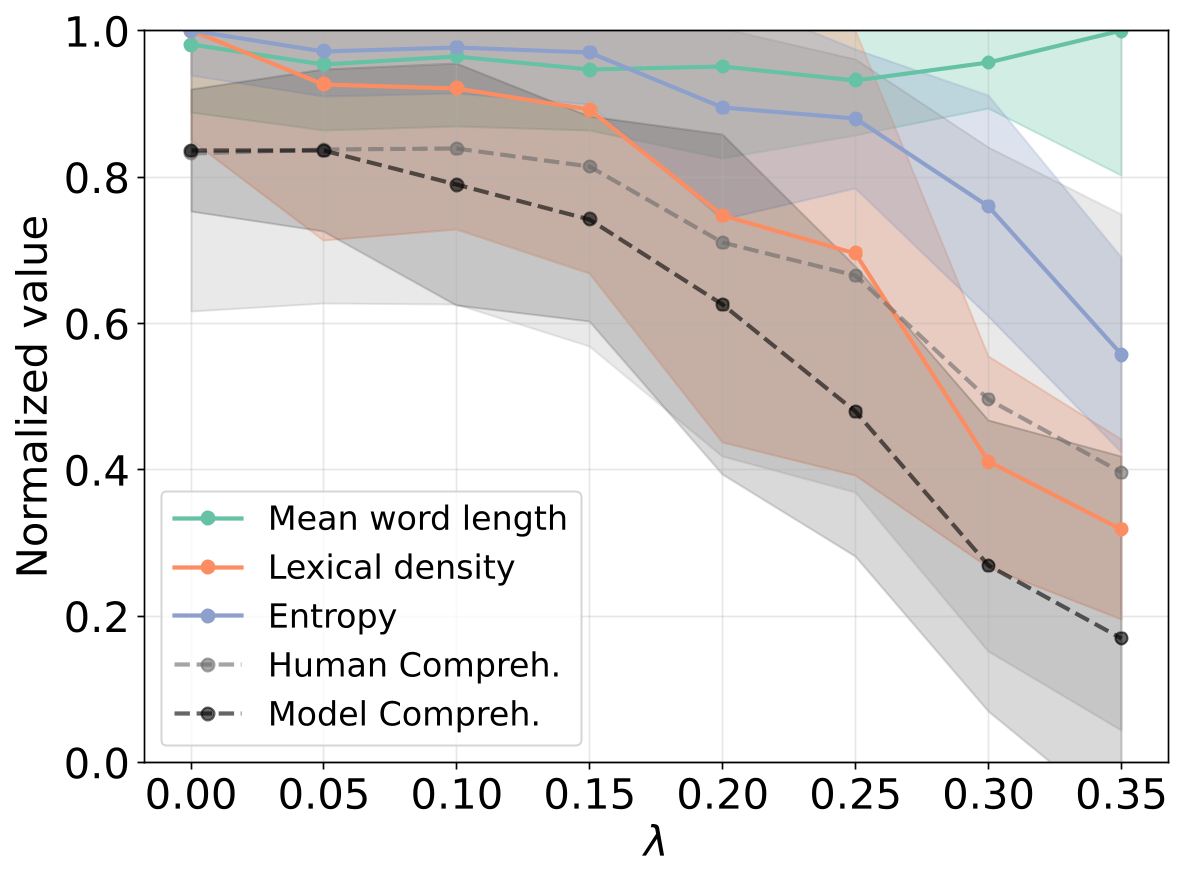}}
\hfill
\subfloat[Fear]{\includegraphics[width=0.3\textwidth]{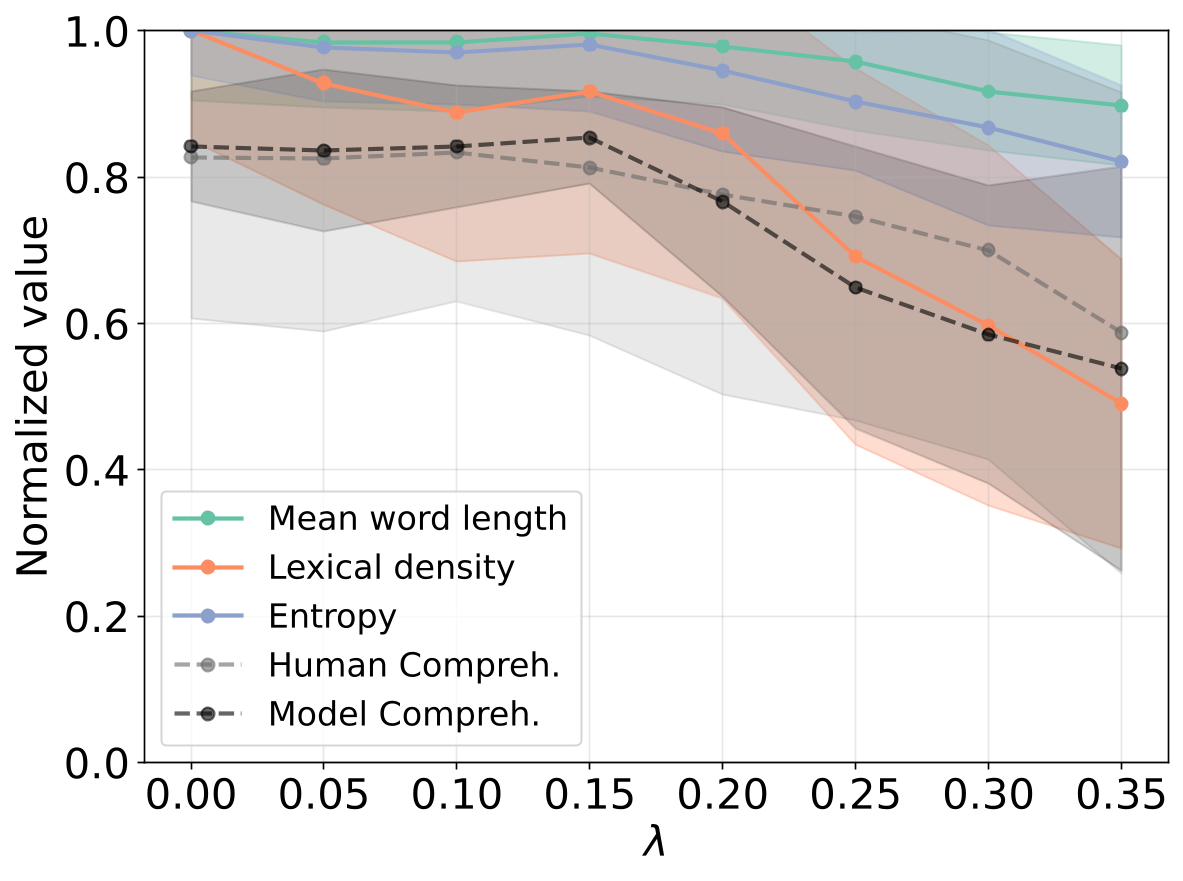}}

\subfloat[Joy]{\includegraphics[width=0.3\textwidth]{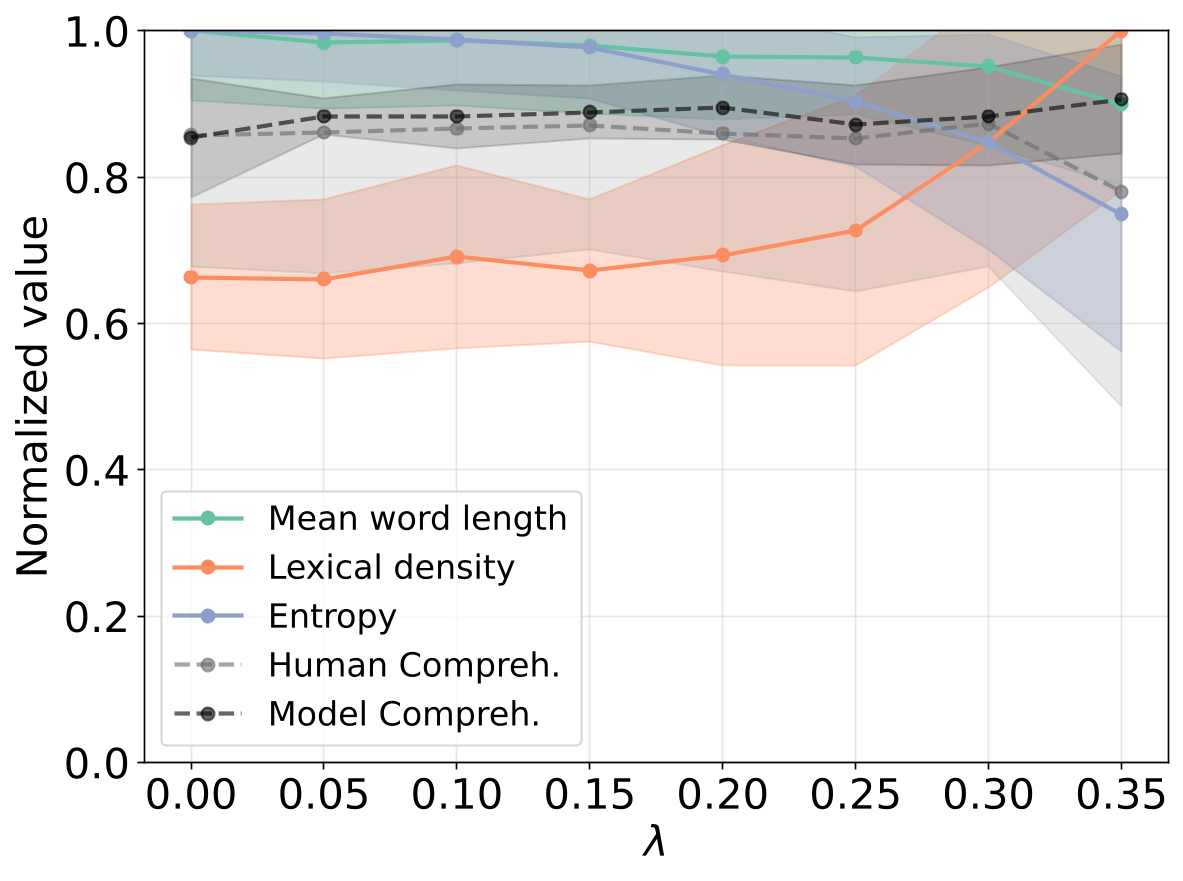}}
\hfill
\subfloat[Sadness]{\includegraphics[width=0.3\textwidth]{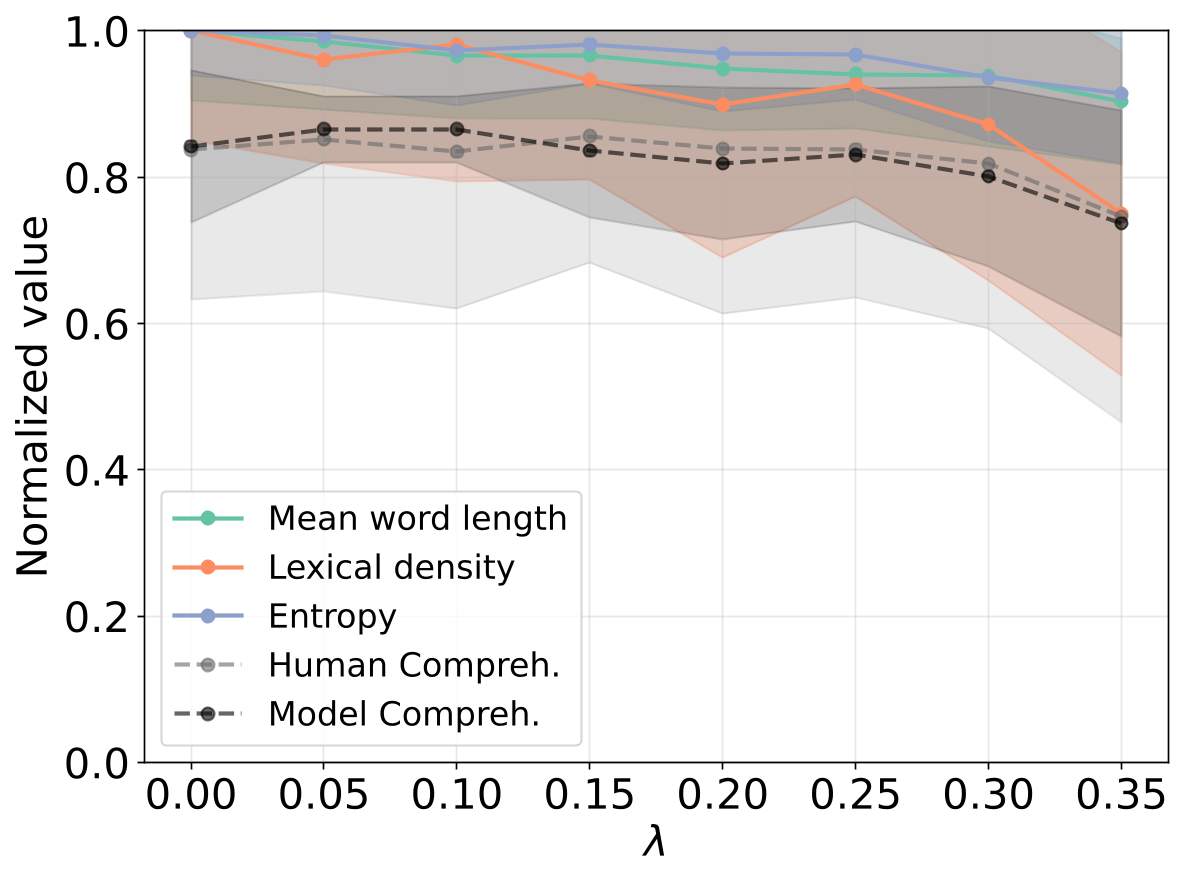}}
\hfill
\subfloat[Surprise]{\includegraphics[width=0.3\textwidth]{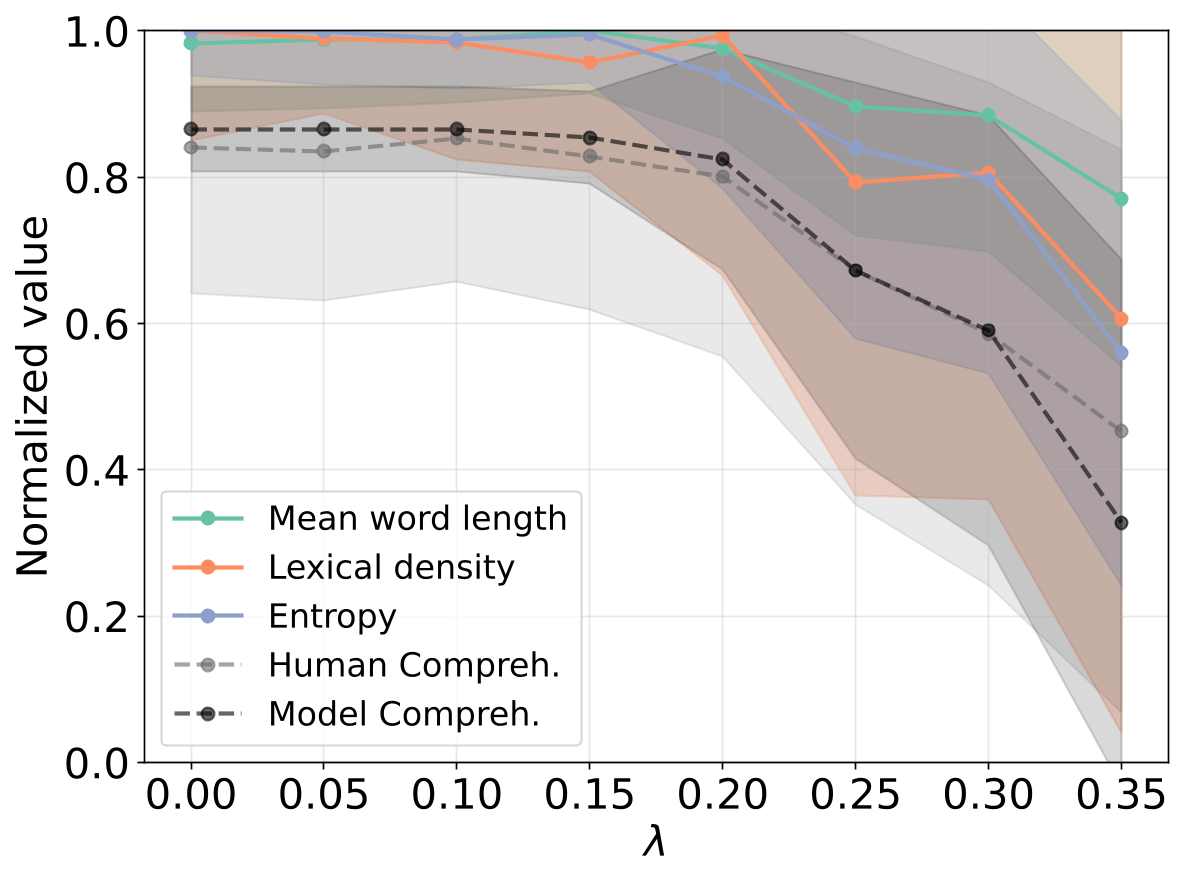}}

\caption{Normalized text features and comprehensibility scores across steering strengths \(\lambda\). Each plot corresponds to one target emotion and shows five normalized values: lexical density, mean word length, entropy, and comprehensibility as rated by humans and a model. All values are scaled to their respective maximum to illustrate relative trends.}

\label{fig:text_feature_comparison}
\end{figure*}

\subsection{Impact on Text Quality}

A core question for activation‑based steering is whether increasing the steering strength (\(\lambda\)) pushes the model beyond regions of its latent space that support coherent language generation. Because style vectors perturb internal activations—especially at higher magnitudes—there is a genuine risk of degrading semantic and syntactic structure. Prior work has shown that substantial activation edits can yield incoherent, repetitive, or semantically inconsistent outputs~\cite{rimsky2024steering, dathathri2019plug, yang2021fudge, silva2025steering}. We therefore examined how text quality evolves with \(\lambda\), combining direct comprehensibility ratings with surface‑level text features.

Fig.~\ref{fig:text_feature_comparison} summarizes five normalized 
measures across \(\lambda\): mean word length, lexical density, entropy, and both human and model comprehensibility scores.  
\emph{Mean word length} serves as a rough proxy for lexical sophistication.  
\emph{Lexical density} is the proportion of content words (nouns, verbs, adjectives, adverbs) in a sentence and signals information richness.  
\emph{Entropy} captures vocabulary diversity: it is the Shannon entropy of the word‑frequency distribution in the text, so higher values indicate more varied word use, whereas low values reflect repetition.

For most emotions—\emph{disgust}, \emph{fear}, \emph{anger}, and \emph{surprise}—all text features remained relatively stable up to ($\lambda\!\approx\!0.15$--$0.2$), after which they began to decline. The steepest drops were observed for \emph{disgust} and \emph{surprise}, followed by \emph{anger} and \emph{fear}, while \emph{sadness} and \emph{joy} exhibited only a slight decrease at the highest \(\lambda\). In contrast to all other emotions, lexical density for \emph{joy} rises, starting at around \(\lambda=0.25\) and approaching its maximum at \(\lambda=0.35\). A plausible explanation is that joy functions as a baseline affect in many unsteered generations; steering may therefore reinforce lexical patterns already typical of the model.

Human and model comprehensibility curves align closely—both begin to decline around the same \(\lambda\) threshold, and the relative order across emotions is consistent. Among the surface features, lexical density shows the strongest correspondence with these curves, followed by entropy, while mean word length appears only weakly related.

The standard deviations of the human ratings highlight two critical observations: first, human perceptions of comprehensibility vary considerably, especially at higher steering levels; second, the variability suggests that the comprehensibility of generated text is influenced not only by the intensity of steering but also by the specific prompt used. In contrast, the model outputs show markedly lower variability, likely reflecting its more consistent—though less nuanced—assessment of text quality across conditions.

However, it is essential to acknowledge that comprehensibility can vary considerably across participants. Given the simple rating instruction (see Section~\ref{subsubsec:text_comprehensibility}), human raters may have interpreted the concept differently: some may have focused on surface-level language quality—such as grammatical correctness, fluency, and the absence of awkward repetitions—while others may have evaluated comprehensibility based on content complexity, penalizing texts that were conceptually dense or abstract. Factors like educational background, proficiency in English, and familiarity with the task further influence such individual differences. These subjective variations must be considered when interpreting the human ratings. 

In contrast, the model’s comprehensibility evaluation was based on a fixed prompt explicitly defining the rating criteria: ``This evaluates the clarity and logical coherence of the sentence's content. A score of 1 means the sentence is easy to understand and logically consistent. In contrast, a score of 10 indicates that the sentence is confusing, illogical, or nonsensical (e.g., `an angry cup of coffee').'' While this ensures consistency in model-based evaluation, it may not fully align with the diverse interpretations held by human raters.

Steering intensities up to \(\lambda\!\approx\!0.15\) preserve surface form and clarity, but stronger interventions progressively degrade text quality. Steering parameters, therefore, need emotion‑specific calibration to balance effectiveness and fluency. Moreover, as shown in our earlier study and preliminary experiments, the effect of a given \(\lambda\)-value also depends on the number of layers being steered—meaning that optimal steering strength is shaped by multiple interacting factors, including the model architecture and intervention scope.

\section{Limitations}
\label{sec:limitations}

Our results were obtained with a single model family and a fixed steering configuration (injection at all layers with a shared $\lambda$). Effects may differ for other architectures, instruction-tuned models, or alternative layer-selection schedules. However, in our prior work~\cite{konen2024style} we also examined steering with only three layers and conducted a probing study across many different layers, finding that style-relevant information is robustly encoded from layer~3 onward and particularly strongly in layers~18–20. We therefore expect similar results across reasonable layer choices, provided that $\lambda$ is carefully tuned.

Style vectors were derived from the GoEmotions dataset, which has imbalanced category frequencies. We did not reweight categories during vector construction, so the resulting vectors may partly reflect corpus priors. Although GoEmotions is imbalanced, we observed strong steerability for low-frequency classes (e.g., fear) and weak steerability for higher-frequency surprise, suggesting that frequency alone does not determine steerability; future work will examine additional factors that may drive these differences.

Our evaluation uses 19 single-turn prompts and does not test multi-turn dialogues or more complex interactive tasks. The generalization of steering behavior to these settings remains unexamined.

Overall, we believe these limitations do not undermine the main contributions and findings of this work, as the observed effects are consistent with prior results and align with theoretical expectations of activation-space steering.

\section{Conclusion}
\label{sec:conclusion}

This study presents the first systematic human evaluation of activation-based steering in language models, offering new insights into both the effectiveness and the practical boundaries of this emerging technique. Using over 7,000 human ratings from 190 participants, we systematically assessed how increasing steering strength (\(\lambda\)) affects perceived emotional intensity and the comprehensibility of generated text.

Our results confirm that activation steering effectively amplifies target emotion signals in text: human raters consistently perceived increasing intensity for five out of six primary emotions as \(\lambda\) increased. At the same time, we identify emotion-specific thresholds—typically around \(\lambda \approx 0.15\)—beyond which comprehensibility and linguistic coherence begin to degrade. This highlights the need for emotion-sensitive calibration when applying steering vectors in practical settings.

A second key finding concerns evaluation methodology. Across a wide range of prompts and emotions, model-based quality ratings closely mirror the trends observed in human judgments, especially in the decline of text clarity at higher \(\lambda\) values. This alignment suggests that automatic evaluation can serve as a reliable proxy for perceived quality in future studies, substantially reducing the need for costly manual annotation.

Importantly, we show that perceived emotional intensity is not only targetable but also measurable at scale, even when the cues are subtle or embedded in context. These findings validate activation steering as a viable mechanism for lightweight behavioral control in LLMs and provide clear guidance on balancing strength and stability.

Future work should explore the generalizability of these results beyond emotional tone—for example, to more abstract or stylistic forms—and refine automatic metrics that capture not only comprehensibility, but also nuance, tone, and intent. As LLMs integrate into everyday applications, the ability to steer their behavior in fine-grained and interpretable ways will become increasingly important. Our study provides a foundation for making this control both effective and scalable.





\appendices

\section{Prompts}
\label{app:prompts}
\begin{enumerate}
    \item Come up with a fairy tale.
    \item Comment on a review of a business written by a customer. 
    \item Compare the cultural value of theatres and cinemas.
    \item Compare the quality of coffee and tea.
    \item Describe a fictional character.
    \item Describe an idea of what a fulfilling and meaningful life should be like.
    \item Describe a piece of art.
    \item Describe a typical day for a family in the 1980s.
    \item Describe an imaginary board meeting at which the concept of PopTarts is proposed.
    \item Pick a colour. Write about that colour as if it was a person.
    \item Rewrite a famous film scene from the perspective of a fish.
    \item Summarise the plot of ‘The Lord of the Rings’ by J.R.R. Tolkien.
    \item Write a couple of lyrics about society.
    \item Write a letter to your parents.
    \item Write a scene between a mentor and a mentee in which the mentee must stand up for himself or herself because the mentor doesn't give him or her enough credit.
    \item Write a story about an encounter of a boy and a girl.
    \item Write a story that involves magic.
    \item Write a story that starts with `The moon was bright. Not that the person could see it.’.
    \item Write an opinion on the role of government in society.
\end{enumerate}

\section{Example}
\label{app:example}

\paragraph{Example steered text: \textit{Describe a typical day for a family in the 1980s.}}

\subsubsection*{Unsteered}
The 1980s — a time of neon-colored clothing, big hair, and radical music. For a typical family in the 1980s, a day would likely look like this:

Morning:  
The family wakes up early, with the kids rushing to get ready for school. Mom and Dad are already up, making breakfast in the kitchen. The smell of bacon and toast fills the air as they scramble to get everyone out the door on time. The kids grab their backpacks and head out the door, while Mom and Dad exchange a quick kiss and a “have a good day” before sending them off to school.

Afternoon:  
Mom heads to the grocery store to pick up some essentials for dinner. She navigates the aisles, carefully selecting the freshest produce and meats for tonight’s meal. Meanwhile, Dad heads to work, commuting to his office job in a suit and tie. He spends the day typing away on his computer, attending meetings, and making deals.

Evening:  
The family gathers around the dinner table, where Mom has prepared a hearty meal of spaghetti and meatballs. They share stories about their day, from the kids’ adventures at school to Dad’s triumphs at work. After dinner, the family watches TV together, laughing at the antics of \textit{The Cosby Show} or \textit{Family Ties}. They might even play a board game or two, like Monopoly or Scrabble.

Night:  
As the evening winds down, the family heads to bed, exhausted from the day’s activities. Mom and Dad tuck the kids in, giving them a goodnight kiss and a reminder to brush their teeth. As they drift off to sleep, they dream of the next day’s adventures.

\subsubsection*{Steered: 0.2 Anger}
A typical day for a family in the 1980s was vastly different from today. Here's a glimpse into what a typical day might look like:

The family wakes up early, around 6:00 AM. The father, a working-class man, gets up first and starts getting ready for work. He puts on his suit and tie, grabs his briefcase, and heads out the door to catch the bus to work.

The mother, a homemaker, stays behind to get the kids ready for school. She makes breakfast, usually something simple like cereal or toast, and gets the kids dressed and packed off to school.

The kids, two boys and a girl, all under the age of 12, are forced to wear the same clothes every day because their parents can't afford to buy them new ones. They wear hand-me-downs from older siblings or, worse, repurposed adult clothing.

The mother then spends the rest of the day cleaning the house, doing laundry, and cooking dinner. She is forced to do all of this by hand, because she can’t afford a washing machine or a dishwasher.

The father comes home from work at 5:00 PM, exhausted and stressed from dealing with his boss and coworkers all day. He spends the evening watching TV and drinking beer, while the mother continues with housework.

The kids do their homework in front of the TV because they can’t afford a desk or computer. They use outdated textbooks passed down from their parents.

\subsubsection*{Steered: 0.3 Joy}
The 1980s! A time of big hair, neon colors, and lots of fun! Here’s a glimpse into a typical day for a family in the 80s:

Morning:  
The family wakes up to the sound of the alarm clock, which is probably a cassette player blasting a favorite song. Mom and Dad get ready for the day, and the kids help each other get dressed and enjoy a quick breakfast.

Afternoon:  
The family heads out for a fun day! They might go to the park, play games, and soak up the sunshine. The kids enjoy their new toys and the excitement of being outdoors.

Evening:  
After their outing, the family returns home and enjoys a delicious dinner together. They might watch a movie or play a few more games. With snacks and laughter, the family winds down their happy day together.

\subsubsection*{Steered: 0.35 Sadness}
The 1980s — a time of neon-colored clothing, big hair, and a lack of social media. But behind the bright colors was a quiet struggle. Here’s how a typical day might have looked for a struggling family:

It’s Monday morning. The father, a working-class man, wakes early and dresses in worn-out jeans and a faded shirt. He kisses his wife goodbye and leaves for the factory, where he works long hours for little pay.

The mother, a stay-at-home mom, gets the children ready for school. They wear the same old clothes every day because the family can’t afford new ones.

The kids, two boys and a girl, feel sad and distant. They miss their father and don’t understand why he always has to leave. They feel forgotten.

The mother tries to comfort them, but she’s overwhelmed. She spends the day cleaning, worrying, and trying to hold things together.

At school, the children feel isolated. They don’t want to be there, but they don’t want to be home either. They feel unloved in a world that seems unfair.

\begin{landscape}
\section{Descriptive statistics}
\label{app:descriptive_stats}

\begin{table}[h]
\centering
\caption{Means and standard deviations of perceived emotional intensity and text comprehensibility of all steering conditions}
\tiny{
\begin{tabular}{l c c c c c c c c c c c c c c c c c c c c c c c c c c}  
\toprule  
Target emotion & \multicolumn{8}{c}{Anger} &  & \multicolumn{8}{c}{Disgust} &  & \multicolumn{8}{c}{Fear} \\
\cmidrule(lr){2-9} \cmidrule(lr){11-18} \cmidrule(l){20-27}
\(\lambda\) & 0.00 & 0.05 & 0.10 & 0.15 & 0.20 & 0.25 & 0.30 & 0.35 &  & 0.00 & 0.05 & 0.10 & 0.15 & 0.20 & 0.25 & 0.30 & 0.35 &  & 0.00 & 0.05 & 0.10 & 0.15 & 0.20 & 0.25 & 0.30 & 0.35 \\
\cmidrule(lr){2-9} \cmidrule(lr){11-18} \cmidrule(l){20-27}
 & \textit{M} & \textit{M} & \textit{M} & \textit{M} & \textit{M} & \textit{M} & \textit{M} & \textit{M} &  & \textit{M} & \textit{M} & \textit{M} & \textit{M} & \textit{M} & \textit{M} & \textit{M} & \textit{M} &  & \textit{M} & \textit{M} & \textit{M} & \textit{M} & \textit{M} & \textit{M} & \textit{M} & \textit{M} \\
 & \textit{(SD)} & \textit{(SD)} & \textit{(SD)} & \textit{(SD)} & \textit{(SD)} & \textit{(SD)} & \textit{(SD)} & \textit{(SD)} &  & \textit{(SD)} & \textit{(SD)} & \textit{(SD)} & \textit{(SD)} & \textit{(SD)} & \textit{(SD)} & \textit{(SD)} & \textit{(SD)} &  & \textit{(SD)} & \textit{(SD)} & \textit{(SD)} & \textit{(SD)} & \textit{(SD)} & \textit{(SD)} & \textit{(SD)} & \textit{(SD)} \\
Emotional intensity & & & & & & & & & & & & & & & & & & & & & & & & & & \\
\midrule
Anger & 0.91 & 1.11 & 1.97 & 1.63 & 2.34 & 2.95 & 3.36 & 3.10 &  & 1.00 & 1.32 & 1.51 & 1.81 & 2.32 & 2.97 & 3.09 & 2.50 &  & 1.04 & 1.03 & 1.28 & 1.17 & 1.25 & 1.57 & 1.35 & 1.03 \\
 & (1.82) & (2.03) & (2.45) & (2.26) & (2.58) & (2.69) & (2.83) & (2.76) &  & (1.91) & (2.18) & (2.34) & (2.45) & (2.65) & (2.69) & (2.69) & (2.70) &  & (1.90) & (1.95) & (2.08) & (1.98) & (1.93) & (2.20) & (2.10) & (1.78) \\
Disgust & 0.70 & 0.88 & 1.05 & 1.09 & 1.58 & 2.57 & 2.57 & 2.68 &  & 0.75 & 0.78 & 1.32 & 1.98 & 3.21 & 4.54 & 4.60 & 4.86 &  & 0.74 & 0.79 & 0.96 & 1.02 & 0.93 & 1.53 & 1.20 & 1.18 \\
 & (1.55) & (1.81) & (1.92) & (1.95) & (2.21) & (2.60) & (2.66) & (2.75) &  & (1.74) & (1.73) & (2.19) & (2.50) & (2.76) & (2.52) & (2.64) & (2.51) &  & (1.63) & (1.70) & (1.70) & (1.82) & (1.70) & (2.22) & (1.95) & (1.90) \\
Fear & 0.98 & 1.15 & 1.15 & 1.43 & 1.38 & 1.47 & 1.51 & 1.51 &  & 0.93 & 1.13 & 1.76 & 1.48 & 1.61 & 1.76 & 0.98 & 0.92 &  & 0.85 & 1.35 & 2.29 & 2.17 & 2.53 & 4.03 & 4.50 & 4.80 \\
 & (1.69) & (1.93) & (1.84) & (2.09) & (2.06) & (2.15) & (2.30) & (2.17) &  & (1.67) & (1.81) & (2.29) & (2.18) & (2.31) & (2.29) & (1.72) & (1.83) &  & (1.55) & (2.15) & (2.61) & (2.46) & (2.55) & (2.54) & (2.48) & (2.31) \\
Joy & 2.43 & 1.85 & 1.76 & 1.63 & 1.00 & 0.66 & 0.62 & 0.59 &  & 2.49 & 1.74 & 1.51 & 1.01 & 0.44 & 0.26 & 0.16 & 0.06 &  & 2.39 & 1.68 & 1.20 & 1.20 & 1.05 & 0.41 & 0.24 & 0.19 \\
 & (2.50) & (2.26) & (2.13) & (2.28) & (1.86) & (1.58) & (1.57) & (1.37) &  & (2.51) & (2.27) & (2.17) & (1.86) & (1.23) & (0.97) & (0.74) & (0.25) &  & (2.43) & (2.22) & (2.00) & (1.97) & (1.92) & (1.18) & (0.82) & (0.83) \\
Sadness & 1.24 & 1.42 & 1.49 & 1.33 & 1.67 & 1.98 & 1.84 & 1.73 &  & 1.20 & 1.78 & 1.51 & 2.17 & 1.80 & 2.12 & 1.74 & 1.20 &  & 1.18 & 1.65 & 1.65 & 1.69 & 1.91 & 2.17 & 2.28 & 2.32 \\
 & (1.96) & (2.08) & (2.09) & (2.08) & (2.22) & (2.48) & (2.41) & (2.33) &  & (1.97) & (2.28) & (2.21) & (2.42) & (2.43) & (2.48) & (2.40) & (2.01) &  & (1.89) & (2.29) & (2.25) & (2.16) & (2.25) & (2.37) & (2.48) & (2.38) \\
Surprise & 1.56 & 1.73 & 1.41 & 1.38 & 1.40 & 1.25 & 1.07 & 1.19 &  & 1.45 & 1.47 & 1.44 & 1.36 & 1.44 & 1.36 & 1.11 & 0.78 &  & 1.50 & 1.23 & 1.48 & 1.53 & 1.61 & 1.63 & 1.42 & 1.45 \\
 & (2.02) & (2.09) & (2.03) & (2.02) & (2.03) & (1.96) & (1.84) & (1.88) &  & (1.94) & (1.99) & (1.97) & (1.88) & (2.08) & (2.11) & (1.86) & (1.62) &  & (2.01) & (1.85) & (2.06) & (1.99) & (2.08) & (2.15) & (2.09) & (1.98) \\
Comprehensibility & 5.91 & 5.90 & 5.79 & 5.66 & 5.58 & 5.16 & 5.00 & 4.10 &  & 5.83 & 5.86 & 5.87 & 5.70 & 4.97 & 4.66 & 3.47 & 2.77 &  & 5.79 & 5.78 & 5.84 & 5.69 & 5.43 & 5.22 & 4.90 & 4.11 \\
 & (1.33) & (1.39) & (1.51) & (1.68) & (1.62) & (1.91) & (2.02) & (2.33) &  & (1.52) & (1.47) & (1.49) & (1.72) & (2.05) & (2.07) & (2.41) & (2.47) &  & (1.55) & (1.66) & (1.43) & (1.61) & (1.91) & (1.95) & (2.01) & (2.30) \\
\midrule
\midrule
Target emotion & \multicolumn{8}{c}{Joy} &  & \multicolumn{8}{c}{Sadness} &  & \multicolumn{8}{c}{Surprise} \\
\cmidrule(lr){2-9} \cmidrule(lr){11-18} \cmidrule(l){20-27}
\(\lambda\) & 0.00 & 0.05 & 0.10 & 0.15 & 0.20 & 0.25 & 0.30 & 0.35 &  & 0.00 & 0.05 & 0.10 & 0.15 & 0.20 & 0.25 & 0.30 & 0.35 &  & 0.00 & 0.05 & 0.10 & 0.15 & 0.20 & 0.25 & 0.30 & 0.35 \\
\cmidrule(lr){2-9} \cmidrule(lr){11-18} \cmidrule(l){20-27}
 & \textit{M} & \textit{M} & \textit{M} & \textit{M} & \textit{M} & \textit{M} & \textit{M} & \textit{M} &  & \textit{M} & \textit{M} & \textit{M} & \textit{M} & \textit{M} & \textit{M} & \textit{M} & \textit{M} &  & \textit{M} & \textit{M} & \textit{M} & \textit{M} & \textit{M} & \textit{M} & \textit{M} & \textit{M} \\
 & \textit{(SD)} & \textit{(SD)} & \textit{(SD)} & \textit{(SD)} & \textit{(SD)} & \textit{(SD)} & \textit{(SD)} & \textit{(SD)} &  & \textit{(SD)} & \textit{(SD)} & \textit{(SD)} & \textit{(SD)} & \textit{(SD)} & \textit{(SD)} & \textit{(SD)} & \textit{(SD)} &  & \textit{(SD)} & \textit{(SD)} & \textit{(SD)} & \textit{(SD)} & \textit{(SD)} & \textit{(SD)} & \textit{(SD)} & \textit{(SD)} \\
Emotional intensity & & & & & & & & & & & & & & & & & & & & & & & & & & \\
\midrule
Anger & 0.95 & 0.43 & 0.31 & 0.41 & 0.28 & 0.23 & 0.13 & 0.16 &  & 0.96 & 0.84 & 0.85 & 0.94 & 1.08 & 0.92 & 1.00 & 0.87 &  & 1.08 & 0.96 & 0.73 & 0.88 & 0.57 & 0.66 & 0.55 & 0.48 \\
 & (1.87) & (1.08) & (0.92) & (1.06) & (0.87) & (0.92) & (0.60) & (0.64) &  & (1.87) & (1.67) & (1.65) & (1.69) & (1.80) & (1.79) & (1.76) & (1.57) &  & (1.97) & (1.91) & (1.55) & (1.77) & (1.41) & (1.51) & (1.31) & (1.37) \\
Disgust & 0.72 & 0.33 & 0.19 & 0.21 & 0.18 & 0.19 & 0.14 & 0.15 &  & 0.73 & 0.54 & 0.68 & 0.43 & 0.61 & 0.71 & 0.66 & 0.71 &  & 0.76 & 0.53 & 0.56 & 0.50 & 0.56 & 0.54 & 0.41 & 0.44 \\
 & (1.70) & (0.91) & (0.65) & (0.76) & (0.65) & (0.78) & (0.71) & (0.66) &  & (1.64) & (1.33) & (1.58) & (1.06) & (1.26) & (1.53) & (1.43) & (1.45) &  & (1.66) & (1.41) & (1.39) & (1.34) & (1.35) & (1.40) & (1.03) & (1.37) \\
Fear & 1.00 & 0.76 & 0.41 & 0.53 & 0.35 & 0.32 & 0.24 & 0.20 &  & 0.99 & 1.20 & 1.19 & 1.26 & 1.40 & 1.34 & 1.50 & 1.77 &  & 0.91 & 0.98 & 0.93 & 0.99 & 1.08 & 0.73 & 0.64 & 0.62 \\
 & (1.75) & (1.63) & (1.08) & (1.25) & (1.05) & (1.07) & (0.95) & (0.77) &  & (1.74) & (1.77) & (1.92) & (1.83) & (1.99) & (1.94) & (2.12) & (2.24) &  & (1.65) & (1.67) & (1.74) & (1.76) & (1.95) & (1.47) & (1.43) & (1.48) \\
Joy & 2.57 & 3.46 & 3.57 & 3.81 & 4.11 & 4.28 & 4.63 & 3.83 &  & 2.48 & 2.07 & 1.78 & 1.87 & 1.25 & 1.28 & 1.03 & 0.81 &  & 2.54 & 2.34 & 2.46 & 2.04 & 1.82 & 1.71 & 1.50 & 0.90 \\
 & (2.57) & (2.63) & (2.58) & (2.51) & (2.60) & (2.43) & (2.38) & (2.62) &  & (2.58) & (2.31) & (2.27) & (2.26) & (1.88) & (1.92) & (1.81) & (1.56) &  & (2.42) & (2.51) & (2.51) & (2.35) & (2.26) & (2.32) & (2.16) & (1.81) \\
Sadness & 1.12 & 1.15 & 0.65 & 0.71 & 0.45 & 0.44 & 0.36 & 0.28 &  & 1.35 & 1.34 & 1.96 & 2.30 & 2.92 & 3.23 & 3.68 & 3.54 &  & 1.13 & 0.94 & 0.97 & 1.18 & 1.05 & 0.95 & 0.64 & 0.51 \\
 & (1.91) & (1.93) & (1.42) & (1.47) & (1.22) & (1.20) & (1.19) & (0.94) &  & (2.05) & (1.92) & (2.49) & (2.45) & (2.55) & (2.58) & (2.56) & (2.74) &  & (1.83) & (1.68) & (1.68) & (1.83) & (1.77) & (1.67) & (1.44) & (1.26) \\
Surprise & 1.47 & 1.38 & 1.58 & 1.43 & 1.23 & 1.17 & 1.16 & 0.91 &  & 1.56 & 1.46 & 1.22 & 1.48 & 1.27 & 1.29 & 1.01 & 1.25 &  & 1.47 & 1.77 & 1.72 & 1.67 & 1.79 & 1.63 & 1.66 & 1.18 \\
 & (1.95) & (1.96) & (2.12) & (2.02) & (1.84) & (1.80) & (1.95) & (1.80) &  & (2.07) & (1.99) & (1.92) & (1.95) & (1.86) & (1.90) & (1.73) & (1.95) &  & (2.04) & (2.12) & (2.11) & (2.17) & (2.15) & (2.12) & (2.10) & (1.91) \\
Comprehensibility & 6.00 & 6.03 & 6.06 & 6.10 & 6.02 & 5.97 & 6.11 & 5.46 &  & 5.86 & 5.96 & 5.84 & 5.99 & 5.87 & 5.87 & 5.73 & 5.22 &  & 5.88 & 5.85 & 5.97 & 5.80 & 5.60 & 4.71 & 4.10 & 3.17 \\
 & (1.26) & (1.35) & (1.26) & (1.19) & (1.32) & (1.46) & (1.36) & (2.05) &  & (1.43) & (1.45) & (1.50) & (1.21) & (1.58) & (1.42) & (1.58) & (1.97) &  & (1.39) & (1.43) & (1.37) & (1.47) & (1.72) & (2.24) & (2.41) & (2.69) \\
\bottomrule

\end{tabular}
}
\end{table}
\label{tab:descriptive_stats}

\end{landscape}

\newpage

\begin{IEEEbiographynophoto}{Diaoulé Diallo} received the B.A. degree in philosophy-neuroscience-cognition from Otto-von-Guericke-University Magdeburg, and the M.Sc. degree in intelligent systems from Bielefeld University. Since 2021, he has been a researcher with the German Aerospace Center (DLR), and a Ph.D. candidate at the University of Bonn.

His research focuses on complex network science and epidemiology, in particular on risk assessment in temporal networks to support data-driven and machine learning–based intervention strategies during pandemics. His second line of research investigates activation-based steering of large language models, aiming to improve their explainability and controllability through representation-level interventions.
\end{IEEEbiographynophoto}

\begin{IEEEbiographynophoto}{Katharina Dworatzyk} received the B.A. degree in philosophy from Rheinische Friedrich-Wilhelms-University Bonn and the M.Sc. degree in cognitive science from Osnabrück University. Since 2019, she has been a researcher with the German Aerospace Center (DLR).

Her research focuses on evaluating the effectiveness of interventions such as community-building and understanding context-specific user requirements, with the goal of informing the development of human-centered solutions for software engineering education and socio-technical systems. \end{IEEEbiographynophoto}

\begin{IEEEbiographynophoto}{Sophie Jentzsch} received the B.Sc. and M.Sc. degrees in psychology in computer science from Technische Universität Darmstadt. Since 2019, she has been a researcher with the German Aerospace Center (DLR), and a Ph.D. candidate at the University of Darmstadt.

Her research focuses on human-centered artificial intelligence and the societal impact of large language models. In particular, she investigates the role of humor and moral reasoning in human–AI interaction, and contributes to interdisciplinary approaches for understanding and guiding AI behavior. She is actively involved in science communication through podcasts and public outreach formats.
\end{IEEEbiographynophoto}

\begin{IEEEbiographynophoto}{Peer Sch\"utt} received a B.Sc. and M.Sc. degree in computer science from the University of Bonn. Since 2022, he has been a researcher with the German Aerospace Center (DLR), and a Ph.D. candidate at the University of Cologne.

His research focuses on anomaly detection in multispectral imagery, focusing on detecting hazardous substances using remote sensing methods. 
Furthermore, he investigates the application of large language models in diverse research domains and administrative workflows to enhance LLM usability in research institutions.
\end{IEEEbiographynophoto}

\begin{IEEEbiographynophoto}{Sabine Theis} received her B.Sc. from the University of Applied Sciences Aachen/Maastricht and M.Sc. in Information Sciences from the University of Amsterdam. She earned her Ph.D. in industrial engineering and ergonomics from RWTH Aachen University in 2019 with a thesis on ergonomic visualization of personal health data.
She worked at Fraunhofer FKIE, investigating workload in AR/VR environments, and as a postdoctoral researcher at RWTH Aachen University on data visualization projects. Currently, she leads the Human Factors in Software Engineering group at the German Aerospace Center (DLR). Her research focuses on human-AI interaction, and human-centered data visualization systems. 
\end{IEEEbiographynophoto}

\begin{IEEEbiographynophoto}{Tobias Hecking} received the B.Sc. degree in computer science from Westfälische Hochschule and the M.Sc. degree in computer science from the University of Duisburg-Essen. He received his Ph.D. in computer science from the University of Duisburg-Essen in 2016. Since 2020, he has been head of the Intelligent Software Systems group at the German Aerospace Center (DLR), Institute of Software Technology.

His research focuses on theories, methods, and tools of artificial intelligence for the development of data-driven intelligent software systems. The focus here is on methods of network analysis, natural language processing and machine learning. The main areas of application are knowledge-based software systems, analysis of complex processes and intelligent user interfaces.
\end{IEEEbiographynophoto}

\EOD

\end{document}